\documentclass[11pt, logo, twocolumn, copyright, colorlinks=true, allcolors=blue]{nvidiatechreport}

\usepackage[colorlinks=true, allcolors=blue]{hyperref}
\usepackage[capitalize]{cleveref}
\crefname{section}{Sec.}{Secs.}
\Crefname{section}{Section}{Sections}
\crefname{figure}{Fig.}{Figs.}
\Crefname{figure}{Figure}{Figures}
\crefname{table}{Tab.}{Tabs.}
\Crefname{table}{Table}{Tables}

\usepackage[most]{tcolorbox}
\usepackage{listings}
\usepackage{bbm}
\usepackage{titletoc} % Appendix
\usepackage{float}
\usepackage{graphicx}
\usepackage{multirow}
\usepackage{amsmath}
\usepackage{amssymb}
\usepackage{xcolor} 
\usepackage{caption}
\usepackage{algorithm}
\usepackage{algorithmic}
\usepackage{pifont} % c mark and x mark
\usepackage{color, colortbl} % for color table
\usepackage{arydshln} % for dash line in table
\usepackage{tabularx} % for matching width of table
\usepackage{footnote} % for footnote
\usepackage{booktabs} % for table
\definecolor{colorful}{rgb}{0.75, 0.87, 0.98}
\definecolor{colorful2}{rgb}{0.75, 0.93, 0.75}
\newcommand{\cmark}{\ding{51}}%
\newcommand{\xmark}{\ding{55}}%

\usepackage[numbers]{natbib}

\title{Masking Teacher and Reinforcing Student for Distilling Vision-Language Models}

\author{
Byung-Kwan Lee,
Yu-Chiang Frank Wang,
Ryo Hachiuma
}

\correspondingauthor{X}

\begin{abstract}
\textbf{Abstract}\\
Large-scale vision–language models (VLMs) have recently achieved remarkable multimodal understanding, but their massive size makes them impractical for deployment on mobile or edge devices. This raises the need for compact yet capable VLMs that can efficiently learn from powerful large teacher. However, distilling knowledge from large teacher to small student remains challenging due to their large size gap: the student often fails to reproduce the teacher's complex, high-dimensional representations, leading to unstable learning and degraded performance. To address this, we propose \textbf{Masters} (\textbf{Mas}king \textbf{te}acher and \textbf{r}einforcing \textbf{s}tudent), a mask-progressive reinforcement learning (RL) distillation framework. \textbf{Masters} first masks and non-dominant weights of the teacher to reduce unnecessary complexity, then progressively restores the teacher from mask to gradually increase the teacher capacity during training. This strategy allows the student to learn richer representations of teacher in a smooth and stable manner. To further refine knowledge transfer, \textbf{Masters} integrates an offline RL stage with two complementary rewards: an accuracy reward that measures the correctness of the generated responses, and a distillation reward that quantifies the ease of their responses' transferability from teacher to student. Unlike online think–answer RL paradigms that are computationally expensive and generate lengthy responses, our offline RL leverages \textit{pre-generated} responses from masked teachers. These provide rich yet efficient guidance, enabling the students to achieve strong performance without requiring the think–answer process. Extensive experiments across diverse VLM benchmarks demonstrate that \textbf{Masters} outperforms existing compact VLMs and partially surpasses large ones, while being far more efficient. Moreover, gradually increasing the teacher sizes during distillation (e.g., from 14B to 38B) yields smoother convergence and stronger generalization than one-shot distillation (e.g., 38B), revealing a scalable path toward efficient and deployable VLMs.
\end{abstract}

\begin{document}
\twocolumn[{%
\renewcommand\twocolumn[1][]{#1}%
\maketitle
\centering
\vspace{-3mm}
\includegraphics[width=\textwidth]{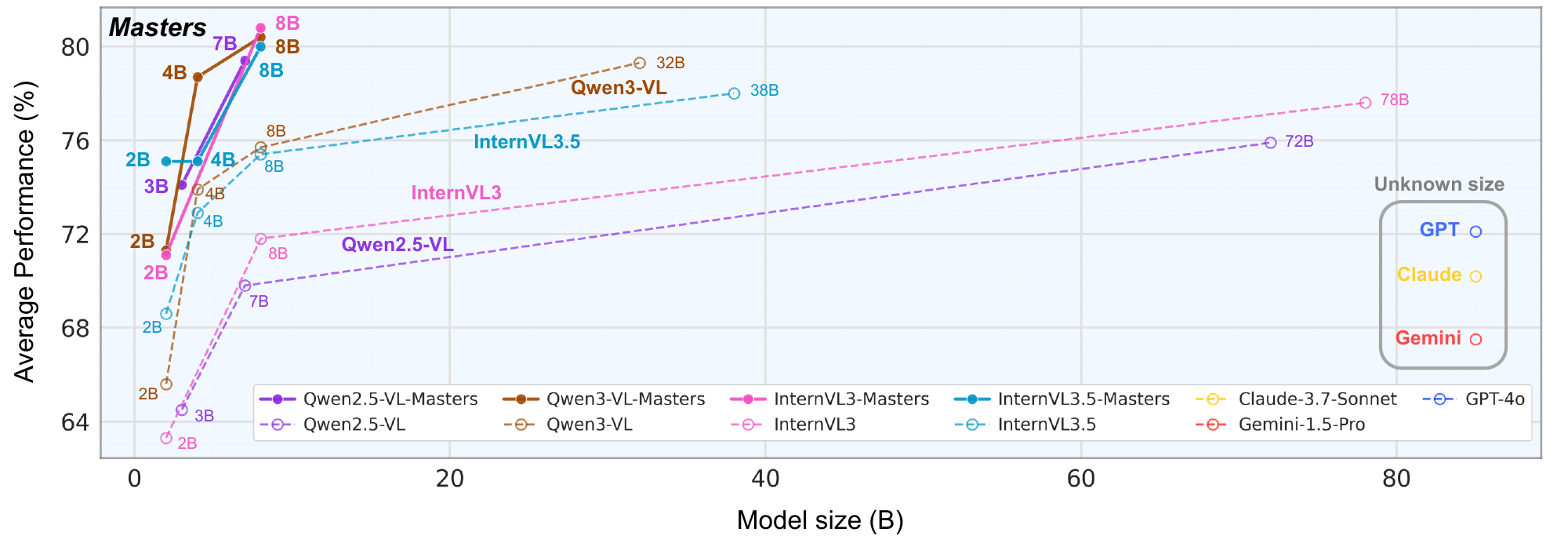}
\vspace{-7mm}
\captionof{figure}{Comparing \textbf{Masters}-applied VLMs with diverse open- and closed-source VLMs across broad model sizes for averaged performance (\%) of numerous evaluation benchmarks: AI2D~\cite{kembhavi2016diagram}, ChartQA~\cite{masry2022chartqa}, MathVista~\cite{lu2023mathvista}, MMB~\cite{liu2023mmbench}, MM-Vet~\cite{yu2023mm}, MMMU~\cite{yue2023mmmu}, MMMU-Pro~\cite{yue2024mmmu}, MMStar~\cite{chen2024we}, BLINK~\cite{fu2024blink}, SEED-Bench~\cite{li2023seed}, SEED-Bench-2-Plus~\cite{li2024seed}, and RealWorldQA.
}
\label{fig:1}
\vspace{5mm}
}]
\begin{figure*}[t!]
    \vspace{0mm}
    \centering
    \includegraphics[width=\textwidth]{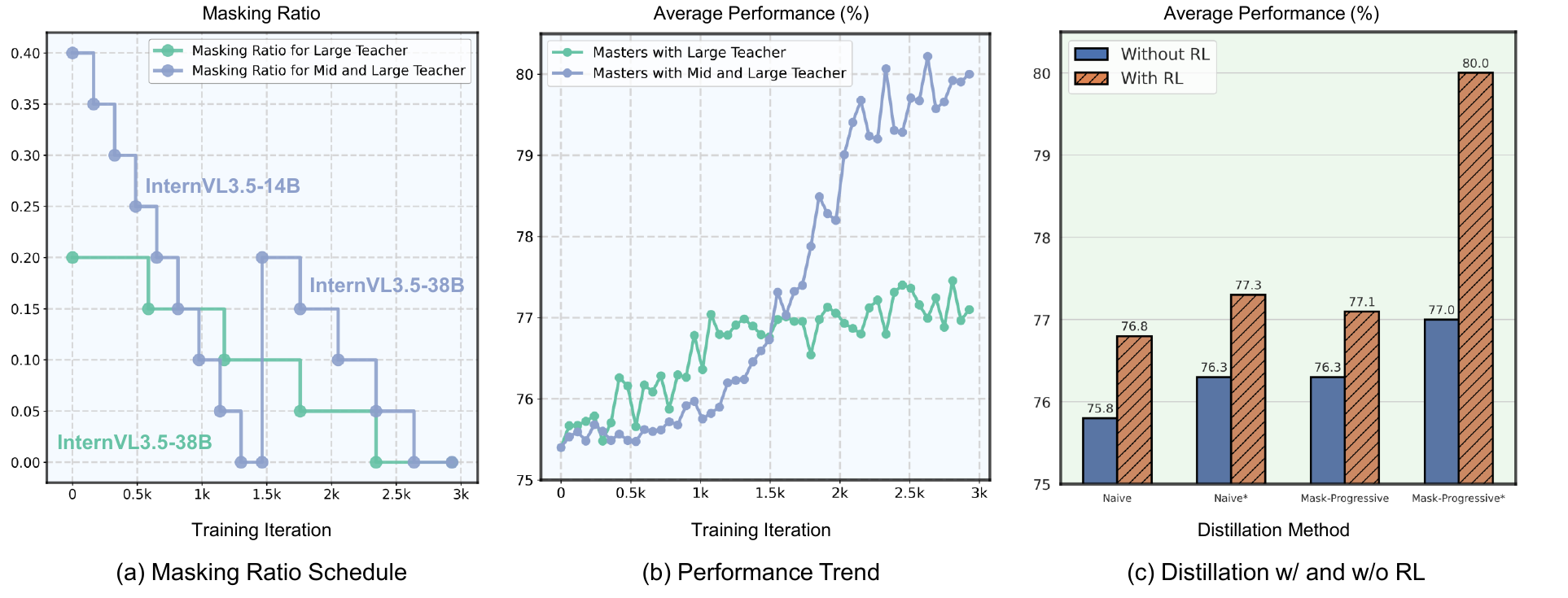}
    \vspace{-7mm}
    \caption{Illustrating training dynamics of \textbf{Masters}, where we represent how (a) mask ratio is controlled during distillation, and (b) its averaged performance log of student (InternVL3.5-8B~\cite{wang2025internvl3}) for evaluation benchmarks in \cref{tab:1}. In addition, we show (c) the effect of RL under naive and mask-progressive distillation. Note that asterisk (*) represents the combined distillation of mid-size and large teacher.}
    \label{fig:2}
    \vspace{0mm}
\end{figure*}
\section{Introduction}
\label{sec:intro}

In recent years, vision–language models (VLMs)~\cite{achiam2023gpt, hurst2024gpt} have demonstrated remarkable capabilities across a wide range of multimodal tasks~\cite{liu2023visual}, including visual captioning, reasoning, and open-ended question answering. By jointly understanding visual and textual information, large-scale VLMs have achieved impressive generalization and reasoning abilities that in some domains approach human-level understanding~\cite{li2025visual}. However, these achievements come at the cost of massive model sizes~\cite{bai2025qwen2, yang2025qwen3, zhu2025internvl3, wang2025internvl3} and heavy computational requirements, making them impractical for deployment on mobile or edge devices~\cite{qu2025mobile}. As the demand for on-device intelligence continues to grow, there is an urgent need for compact yet powerful VLMs~\cite{chu2024mobilevlm, chen2024omnivlm, vasu2025fastvlm, marafioti2025smolvlm} that can deliver competitive performance while maintaining high efficiency and deployability~\cite{sharshar2025vision, yao2025efficient}.

A widely adopted approach for building such lightweight yet capable models is knowledge distillation~\cite{hinton2015distilling, gou2021knowledge}, where a large teacher transfers its knowledge to a smaller student. Despite its promise, distillation remains challenging due to the substantial size gap between teacher and student~\cite{zhang-etal-2023-lifting, guo2020reducing, mirzadeh2020improved}. The student often struggles to reproduce the teacher’s rich and high-dimensional representations, leading to unstable learning and significant performance degradation~\cite{zhou2021bert, yang2022sparse, zhang2023towards}. Recent distillation has explored modified training objectives~\cite{gu2024minillm, xu2024llavadi, ko2024distillm, agarwal2024policy}, dynamic intermediate-layer distillation~\cite{lee2024vlsi, kim2025compodistill}, vision-attention distillation~\cite{feng2024align, cao2025move}, multi-step distillation~\cite{han2024amd, cai2024llava}, cross-token general distillation~\cite{boizard2024towards, lee2025genrecal}, and reinforcement learning (RL)–based approaches~\cite{xu2025kdrl, yang2025multi, lee2025unified}. However, few works have directly addressed the fundamental issue of the large size gap itself.

% figure overview
\begin{figure*}[t!]
    \vspace{0mm}
    \centering
    \includegraphics[width=\textwidth]{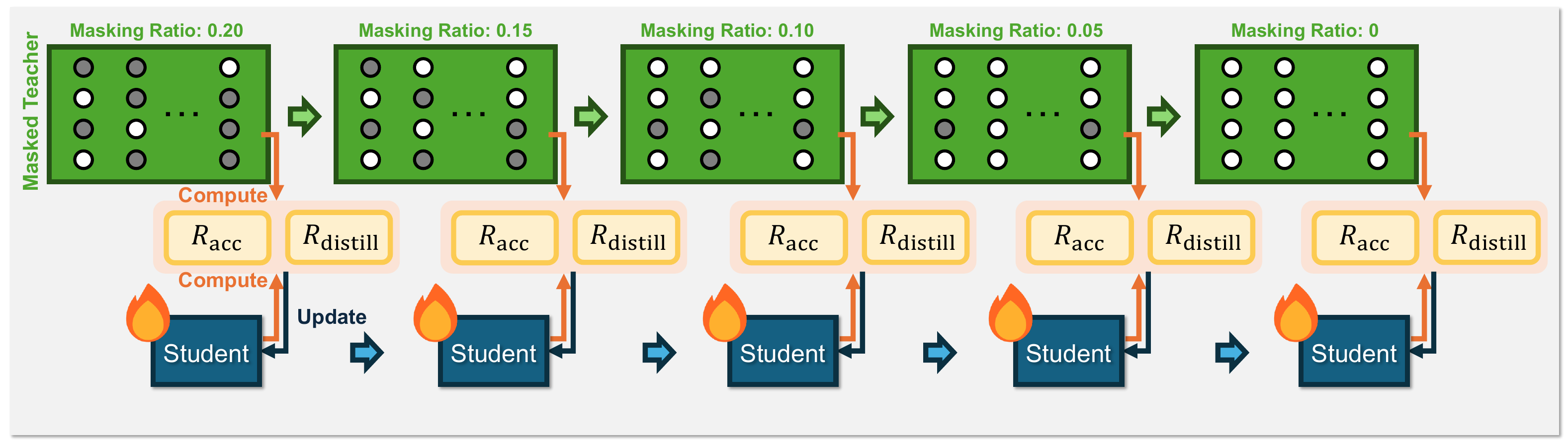}
    \vspace{-7mm}
    \caption{Overview of mask-progressive distillation where teacher is masked with a decreasing masking ratio (0.20, 0.15, 0.10, 0.05, 0), gradually restoring its full capacity. At each masking stage for teacher, student is updated using two rewards: accuracy reward $R_{\text{acc}}$ and distillation reward $R_{\text{distill}}$. This progressive distillation enables smooth and stable knowledge transfer to the student.}
    \label{fig:overview}
    \vspace{0mm}
\end{figure*}

To tackle this challenge, we first mask non-dominant weights of the teacher based on their magnitudes~\cite{han2015learning}, thereby reducing unnecessary complexity. This is because the large number of weights in the teacher is a core factor that makes distillation difficult. After masking the weights in the teacher, we perform the knowledge distillation \textit{progressively} where the teacher is gradually restored from the mask throughout the entire training to increase the teacher’s representation complexity. This mask-progressive strategy enables the student to first capture coarse-grained patterns and subsequently refine them into more detailed and higher-level representations, thereby learning richer and more complex representations of the teacher in a smooth and stable manner.

Furthermore, we identify critical limitations in existing supervised fine-tuning (SFT) datasets~\cite{tong2024cambrian, yue2024mammoth2, an2025llava}, where answer labels are typically generated by large closed-source models with an extremely large number of parameters, such as GPT-4o~\cite{hurst2024gpt}, Gemini~\cite{comanici2025gemini}, or Claude~\cite{claude3series2024}, and often filtered by humans~\cite{xu2024vision}. Small student, however, has limited capacity to learn these rich answer labels due to its smaller vocabulary and lower hidden dimension of representation~\cite{zhao2019extremely, zhang2024dual}. To address this, we utilize the generated responses from masked teachers instead of directly using standard SFT samples. This enables the student to learn the responses that better match the student's current capacity. Moreover, we incorporate the student’s own generated responses into training to maintain alignment between the teacher’s guidance and the student’s evolving representational capacity. This design ensures stable yet continual improvement toward the teacher’s behavior, going beyond the single-answer and over-rich label constraints of conventional SFT datasets.

However, some generated responses may contain factual errors or exhibit linguistic complexity that hinders effective knowledge transfer, ultimately degrading the distillation performance. Hence, we aim to evaluate both the accuracy of the responses and their ease of knowledge transfer, and further refine the student to avoid such undesirable responses. To achieve this, we integrate RL into the distillation process. Specifically, we compute two types of rewards: an accuracy reward evaluated by LLM-as-a-Judge~\cite{zheng2023judging} to account for flexible and diverse answers, and a distillation reward that measures how easily knowledge can be transferred from the teacher to the student. In practice, conventional online RL is too slow and inefficient training, since it requires the model to repeatedly generate multiple responses at every training step under the recent ``think–answer'' paradigm~\cite{guo2025deepseek, shao2024deepseekmath}. As a result, only a very limited amount of data samples~\cite{lu2025ui, peng2025lmm, yu2025perception, liu2025noisyrollout, shen2025vlm, liu2025visual, huang2025vision, lee2025unified} are utilized, constraining the scale and diversity of training samples. To address these limitations, we adopt an offline RL approach in which both the teacher and the student \textit{pre-generate} their multiple responses for all questions, without explicit think-answer. These \textit{pre-generated} responses are then used for RL training, significantly reducing training time and computational costs.

Bringing these components together, we propose a mask-progressive RL distillation framework, referred to as \textbf{Mas}king \textbf{te}acher and \textbf{r}einforcing \textbf{s}tudents (\textbf{Masters}). It integrates teacher weight masking, progressive distillation, multi-response learning, and offline RL into a unified training paradigm. Through this design, it enables the student to effectively absorb knowledge from large teachers while maintaining high efficiency and stability. Through extensive experiments across diverse evaluation benchmarks, we demonstrate that \textbf{Masters} consistently outperforms existing compact VLMs and exceeds large ones in \cref{fig:1}. Beyond empirical gains, we find that gradually increasing the teacher sizes during distillation (e.g., from 14B to 38B) leads to smoother convergence and superior generalization compared to one-shot distillation from a single large teacher (e.g., 38B), revealing an effective pathway for scalable model compression. We believe this framework marks an important step toward efficient, deployable, and continually improvable VLMs for on-device intelligence.

Our main contributions can be summarized in threefold as follows:

% figure 3
\begin{figure*}[t!]
    \vspace{0mm}
    \centering
    \includegraphics[width=\textwidth]{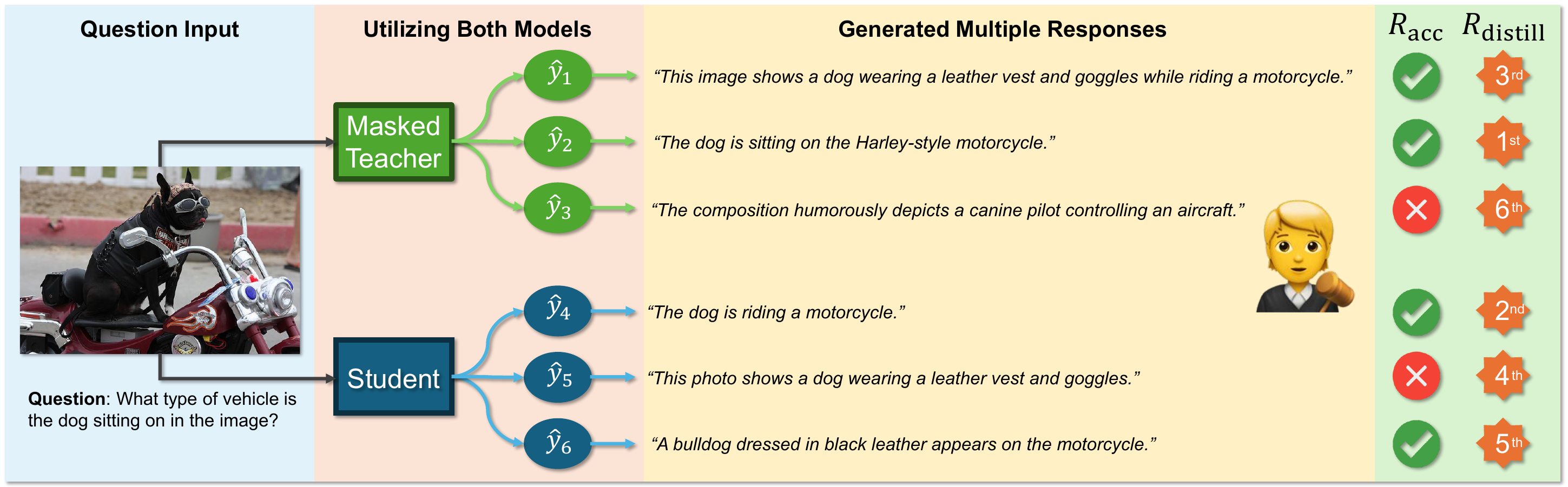}
    \vspace{-3mm}
    \caption{Depicting multiple responses generated by both the masked teacher and the student, where an \textit{accuracy reward} ($R_{\text{acc}}$) evaluates the binary correctness of each response, and a \textit{distillation reward} ($R_{\text{distill}}$) measures the ease of knowledge transfer based on divergence objective between teacher and student logits. Note that the rank labels (1st, 2nd, 3rd, etc.) in $R_{\text{distill}}$ indicate the relative magnitude of the divergence values, where the smallest divergence (1st) receives the highest reward (1.0) and the largest divergence the lowest reward (0.0).}
    \label{fig:3}
    \vspace{0mm}
\end{figure*}

\begin{itemize}
    \item \textbf{Progressive capacity-aligned distillation:} We propose a progressive teacher–student alignment strategy that adaptively matches the student’s learning capacity through restoring the teacher from the mask.
    
    \item \textbf{Offline RL with dual rewards:} By pre-generating diverse responses from both the teacher and the student, our offline RL substantially reduces training time while efficiently achieving strong performance with two complementary objectives — an \emph{accuracy reward} for correctness and a \emph{distillation reward} for transferability.
    
    \item \textbf{Unified and scalable framework:} We integrate progressive capacity-aligned distillation and offline RL into a unified training framework with scaled-up data, enabling compact VLMs to attain large model-level performance for practical efficiency and deployability.
\end{itemize}
\section{Related Work}
\label{sec:related}

Previous approaches to building efficient VLMs (see Appendix~\ref{sec:appA}) have focused on inserting additional modules, modifying internal architectures, or altering propagation strategies within the models themselves. Recent emerging line of research on efficient VLMs has begun to leverage knowledge distillation~\cite{hinton2015distilling, gou2021knowledge}, where the knowledge of a large, high-capacity teacher is transferred into a smaller, more efficient student. This is because training a small model alone inherently limits its representational capacity, prompting a shift toward utilizing rich logit or intermediate representations from the teacher.

% table 1
\begin{table*}[t!]
\vspace{0mm}
\centering
\caption{Comparing the performance between naive, mask-progressive, and RL-applied distillation on AI2D~\cite{kembhavi2016diagram}, ChartQA~\cite{masry2022chartqa}, MathVista~\cite{lu2023mathvista}, MMB/MMB$^\text{CN}$ ~\cite{liu2023mmbench}, MM-Vet~\cite{yu2023mm}, MMMU~\cite{yue2023mmmu}, MMMU-Pro~\cite{yue2024mmmu}, MMStar~\cite{chen2024we}, BLINK~\cite{fu2024blink}, SEED~\cite{li2023seed}, SEED2+~\cite{li2024seed}, RealWorldQA (RWQA). Note that `$+$Large Teacher' refers to the largest model within the same model family in \cref{sec:4.1}.}
\label{tab:1}
\vspace{-3mm}
\resizebox{\linewidth}{!}{
\renewcommand{\tabcolsep}{1mm}
\begin{tabular}{lccccccccccccccc}
\toprule
VLMs
& AI2D 
& ChartQA
& MathVista 
& MMB 
& MMB$^\text{CN}$ 
& MM-Vet 
& MMMU 
& MMMU-Pro 
& MMStar 
& BLINK
& SEED
& SEED2+
& RWQA
& Avg \\
\midrule
Qwen2.5-VL-7B
& 83.9%AI2D 
& 87.3%ChartQA 
& 67.8%MathVista 
& 83.5%MMB 
& 83.4%MMB-CN
& 71.8%MM-Vet 
& 55.0%MMMU 
& 38.3%MMMU-Pro 
& 63.9%MMStar 
& 56.4%BLINK 
& 77.0%SEED
& 70.4%SEED-2-Plus 
& 68.5 %RWQA
& 69.8\\ %Avg
\cdashline{1-15}\noalign{\vskip 0.5ex}
$+$Large Teacher% Method Name
& 84.3%AI2D 
& 87.9%ChartQA 
& 68.4%MathVista 
& 83.9%MMB 
& 83.8%MMB-CN
& 72.3%MM-Vet 
& 55.6%MMMU 
& 38.9%MMMU-Pro 
& 64.5%MMStar 
& 56.9%BLINK 
& 77.4%SEED
& 70.9%SEED-2-Plus 
& 68.9%RWQA
& 70.3\\ %Avg
$+$Mask-Progressive% Method Name
& 85.1%AI2D 
& 89.2%ChartQA 
& 70.3%MathVista 
& 84.8%MMB 
& 84.6%MMB-CN
& 74.1%MM-Vet 
& 57.0%MMMU 
& 40.2%MMMU-Pro 
& 66.2%MMStar 
& 58.5%BLINK 
& 78.5%SEED
& 71.9%SEED-2-Plus 
& 70.3%RWQA
& 71.6\\ %Avg
\rowcolor{colorful}
$+$Reward Feedback% Method Name
& \textbf{86.3}%AI2D 
& \textbf{92.8}%ChartQA 
& \textbf{73.4}%MathVista 
& \textbf{86.7}%MMB 
& \textbf{86.5}%MMB-CN
& \textbf{77.0}%MM-Vet 
& \textbf{59.8}%MMMU 
& \textbf{42.9}%MMMU-Pro 
& \textbf{69.0}%MMStar 
& \textbf{61.3}%BLINK 
& \textbf{80.2}%SEED
& \textbf{73.8}%SEED-2-Plus 
& \textbf{72.4}%RWQA
& \textbf{74.0}\\ %Avg
\midrule
Qwen3-VL-8B % Method Name
& 85.7%AI2D 
& 88.4%ChartQA 
& 77.2%MathVista 
& 86.8%MMB 
& 86.5%MMB-CN
& 74.5%MM-Vet 
& 69.6%MMMU 
& 55.9%MMMU-Pro 
& 70.9%MMStar 
& 69.1%BLINK 
& 78.5%SEED
& 70.8%SEED-2-Plus 
& 69.9%RWQA
& 75.7\\ %Avg
\cdashline{1-15}\noalign{\vskip 0.5ex}
$+$Large Teacher% Method Name
& 86.4%AI2D 
& 90.3%ChartQA 
& 78.4%MathVista 
& 87.5%MMB 
& 87.4%MMB-CN
& 75.8%MM-Vet 
& 70.4%MMMU 
& 57.3%MMMU-Pro 
& 73.1%MMStar 
& 69.9%BLINK 
& 79.3%SEED
& 71.9%SEED-2-Plus 
& 71.8%RWQA
& 76.9\\ %Avg
$+$Mask-Progressive% Method Name
& 87.3%AI2D 
& 92.7%ChartQA 
& 79.8%MathVista 
& 88.4%MMB 
& 88.5%MMB-CN
& 77.3%MM-Vet 
& 71.5%MMMU 
& 59.1%MMMU-Pro 
& 76.0%MMStar 
& 70.9%BLINK 
& 80.3%SEED
& 73.3%SEED-2-Plus 
& 74.3%RWQA
& 78.4\\ %Avg
\rowcolor{colorful}
$+$Reward Feedback% Method Name
& \textbf{88.5}%AI2D 
& \textbf{95.9}%ChartQA 
& \textbf{81.8}%MathVista 
& \textbf{89.5}%MMB 
& \textbf{89.9}%MMB-CN
& \textbf{79.4}%MM-Vet 
& \textbf{72.9}%MMMU 
& \textbf{61.4}%MMMU-Pro 
& \textbf{79.7}%MMStar 
& \textbf{72.3}%BLINK 
& \textbf{81.7}%SEED
& \textbf{75.1}%SEED-2-Plus 
& \textbf{77.5} %RWQA
& \textbf{80.4}\\ %Avg
\midrule
InternVL3-8B
& 85.2%AI2D 
& 86.6%ChartQA 
& 71.6%MathVista 
& 83.4%MMB 
& 82.2%MMB-CN
& 78.5%MM-Vet 
& 62.7%MMMU 
& 41.3%MMMU-Pro 
& 68.2%MMStar 
& 55.5%BLINK 
& 77.1%SEED
& 69.7%SEED-2-Plus 
& 70.8%RWQA
& 71.8\\ %Avg
\cdashline{1-15}\noalign{\vskip 0.5ex}
$+$Large Teacher% Method Name
& 85.8%AI2D 
& 87.4%ChartQA 
& 72.2%MathVista 
& 84.0%MMB 
& 82.8%MMB-CN
& 79.1%MM-Vet 
& 63.6%MMMU 
& 42.1%MMMU-Pro 
& 68.9%MMStar 
& 56.2%BLINK 
& 77.7%SEED
& 70.4%SEED-2-Plus 
& 71.5%RWQA
& 72.4\\ %Avg
$+$Mask-Progressive% Method Name
& 86.5%AI2D 
& 88.5%ChartQA 
& 73.0%MathVista 
& 84.8%MMB 
& 83.6%MMB-CN
& 79.8%MM-Vet 
& 64.7%MMMU 
& 43.0%MMMU-Pro 
& 70.1%MMStar 
& 57.4%BLINK 
& 78.5%SEED
& 71.2%SEED-2-Plus 
& 72.8%RWQA
& 73.4\\ %Avg
\rowcolor{colorful}
$+$Reward Feedback% Method Name
& \textbf{88.1}%AI2D 
& \textbf{91.9}%ChartQA 
& \textbf{76.}3%MathVista 
& \textbf{87.3}%MMB 
& \textbf{86.1}%MMB-CN
& \textbf{82.2}%MM-Vet 
& \textbf{67.1}%MMMU 
& \textbf{46.5}%MMMU-Pro 
& \textbf{74.1}%MMStar 
& \textbf{61.0}%BLINK 
& \textbf{80.5}%SEED
& \textbf{73.4}%SEED-2-Plus 
& \textbf{75.3}%RWQA
& \textbf{76.1}\\ %Avg
\midrule
InternVL3.5-8B % Method Name
& 84.0%AI2D 
& 86.7%ChartQA 
& 78.4%MathVista 
& 86.5%MMB 
& 85.9%MMB-CN
& 83.1%MM-Vet 
& 73.4%MMMU 
& 57.2%MMMU-Pro 
& 69.3 %MMStar 
& 59.5 %BLINK 
& 77.4%SEED
& 70.8%SEED-2-Plus 
& 67.5%RWQA
& 75.4\\ %Avg
\cdashline{1-15}\noalign{\vskip 0.5ex}
$+$Large Teacher% Method Name
& 84.8%AI2D 
& 87.3%ChartQA 
& 78.9%MathVista 
& 86.9%MMB 
& 86.3%MMB-CN 
& 83.5%MM-Vet 
& 73.8%MMMU 
& 57.6%MMMU-Pro 
& 69.8%MMStar 
& 60.0%BLINK 
& 77.8%SEED 
& 71.2%SEED-2-Plus 
& 67.9%RWQA 
& 75.8\\ %Avg
$+$Mask-Progressive% Method Name
& 85.3%AI2D 
& 87.8%ChartQA 
& 79.4%MathVista 
& 87.4%MMB 
& 86.8%MMB-CN
& 84.0%MM-Vet 
& 74.3%MMMU 
& 58.1%MMMU-Pro 
& 70.3%MMStar 
& 60.5%BLINK 
& 78.3%SEED
& 71.7%SEED-2-Plus 
& 68.4%RWQA
& 76.3\\ %Avg
\rowcolor{colorful}
$+$Reward Feedback% Method Name
& \textbf{86.1}%AI2D 
& \textbf{88.6}%ChartQA 
& \textbf{80.2}%MathVista 
& \textbf{88.2}%MMB 
& \textbf{87.6}%MMB-CN
& \textbf{84.8}%MM-Vet 
& \textbf{75.1}%MMMU 
& \textbf{58.9}%MMMU-Pro 
& \textbf{71.1}%MMStar 
& \textbf{61.3}%BLINK 
& \textbf{79.1}%SEED
& \textbf{72.5}%SEED-2-Plus 
& \textbf{69.2}%RWQA
& \textbf{77.1}\\ %Avg
\bottomrule
\end{tabular}
}
\vspace{0mm}
\end{table*}

Beyond the scope of VLM research, early studies on knowledge distillation primarily focused on aligning the logits between the teacher and student~\cite{sanh2019distilbert, turc2019well}. Subsequent works extended this paradigm to intermediate feature representations~\cite{romero2014fitnets, sun2019patient, chen2021distilling, ben2022s, wang2024crosskd}, and later introduced probabilistic~\cite{passalis2020probabilistic}, relational~\cite{park2019relational}, and contrastive formulations~\cite{tian2019contrastive} to better capture internal dependencies~\cite{zagoruyko2016paying, yim2017gift, tung2019similarity, heo2019comprehensive}. Several studies have explored multi-teacher frameworks~\cite{timiryasov2023baby, lee2023ensemble, wan2024fusechat, xu2020feature, son2021densely, wang2021distilling} to integrate diverse knowledge sources, while others have leveraged step-by-step reasoning traces from large models to enhance the student’s reasoning and compositional capabilities~\cite{hsieh2023distilling, tian2024beyond}. More recent research has investigated alternative training objectives~\cite{gu2024minillm, xu2024llavadi, ko2024distillm, agarwal2024policy}, dynamic intermediate-layer distillation~\cite{lee2024vlsi, kim2025compodistill}, vision-attention distillation~\cite{feng2024align, cao2025move}, multi-step distillation~\cite{han2024amd, cai2024llava}, cross-token general distillation~\cite{boizard2024towards, lee2025genrecal}, and reinforcement learning (RL)–based approaches~\cite{xu2025kdrl, yang2025multi, lee2025unified}.

Despite these advances, model distillation remains challenging due to the substantial parameter gap between teacher and student~\cite{zhang-etal-2023-lifting, guo2020reducing, mirzadeh2020improved}. The student often struggles to reproduce the teacher’s rich, high-dimensional representations, leading to unstable training and noticeable performance degradation~\cite{zhou2021bert, yang2022sparse, zhang2023towards}. However, only limited efforts have been made to directly narrow the parameter gap between teacher and student.

Unlike existing approaches that primarily focus on modifying distillation objectives, aligning intermediate features, or combining multiple strategies, we aim to directly address the large parameter gap between teacher and student. To this end, we first mask non-dominant weights in the teacher based on their magnitudes~\cite{han2015learning}, effectively reducing the number of active parameters and simplifying the teacher’s representational space. This masking step not only narrows the capacity gap between teacher and student but also filters out noisy or over-parameterized components that hinder stable knowledge transfer. As training proceeds, we progressively restore the teacher from mask, allowing the student to gradually learn complex representations of teacher in a capacity-aligned manner. This leads to smoother optimization and mitigates the instability in direct distillation from large teacher.

% table 2
\begin{table*}[t!]
\vspace{0mm}
\centering
\caption{Comparison of the performance among naive (using large and mid teacher), mask-progressive, and RL-applied distillation. Note that `$+$Mid Teacher' denotes all intermediate models (e.g., 4B, 8B, and 14B) between student (e.g., 2B) and `Large Teacher' (e.g., 38B).}
\label{tab:2}
\vspace{-3mm}
\resizebox{\linewidth}{!}{
\renewcommand{\tabcolsep}{1mm}
\begin{tabular}{lccccccccccccccc}
\toprule
VLMs
& AI2D 
& ChartQA
& MathVista 
& MMB 
& MMB$^\text{CN}$ 
& MM-Vet 
& MMMU 
& MMMU-Pro 
& MMStar 
& BLINK
& SEED
& SEED2+
& RWQA
& Avg \\
\midrule
InternVL3.5-8B % Method Name
& 84.0%AI2D 
& 86.7%ChartQA 
& 78.4%MathVista 
& 86.5%MMB 
& 85.9%MMB-CN
& 83.1%MM-Vet 
& 73.4%MMMU 
& 57.2%MMMU-Pro 
& 69.3 %MMStar 
& 59.5 %BLINK 
& 77.4%SEED
& 70.8%SEED-2-Plus 
& 67.5%RWQA
& 75.4\\ %Avg
\cdashline{1-15}\noalign{\vskip 0.5ex}
$+$Large Teacher% Method Name
& 84.8%AI2D 
& 87.3%ChartQA 
& 78.9%MathVista 
& 86.9%MMB 
& 86.3%MMB-CN 
& 83.5%MM-Vet 
& 73.8%MMMU 
& 57.6%MMMU-Pro 
& 69.8%MMStar 
& 60.0%BLINK 
& 77.8%SEED 
& 71.2%SEED-2-Plus 
& 67.9%RWQA 
& 75.8\\ %Avg
$+$Mid Teacher% Method Name
& 85.4%AI2D 
& 87.9%ChartQA 
& 79.4%MathVista 
& 87.3%MMB 
& 86.8%MMB-CN 
& 84.0%MM-Vet 
& 74.3%MMMU 
& 58.0%MMMU-Pro 
& 70.3%MMStar 
& 60.5%BLINK 
& 78.3%SEED 
& 71.7%SEED-2-Plus 
& 68.4%RWQA 
& 76.3\\ %Avg
$+$Mask-Progressive% Method Name
& 86.0%AI2D 
& 88.8%ChartQA 
& 80.1%MathVista 
& 87.9%MMB 
& 87.4%MMB-CN 
& 84.6%MM-Vet 
& \textbf{74.9}%MMMU 
& \textbf{58.6}%MMMU-Pro 
& 71.0%MMStar 
& 61.2%BLINK 
& 79.0%SEED 
& 72.4%SEED-2-Plus 
& 69.1%RWQA 
& 77.0\\ %Avg
\rowcolor{colorful}
$+$Reward Feedback% Method Name
& \textbf{87.2}%AI2D 
& \textbf{95.1}%ChartQA 
& \textbf{85.0}%MathVista 
& \textbf{88.2}%MMB 
& \textbf{88.3}%MMB-CN
& \textbf{85.6}%MM-Vet 
& 72.7%MMMU 
& 58.1%MMMU-Pro 
& \textbf{80.8}%MMStar 
& \textbf{67.8}%BLINK 
& \textbf{81.4}%SEED
& \textbf{75.5}%SEED-2-Plus 
& \textbf{74.9} %RWQA
& \textbf{80.0}\\ %Avg
\rowcolor{colorful2}
$-$Mid Teacher% Method Name
& 86.1%AI2D 
& 88.6%ChartQA 
& 80.2%MathVista 
& 88.2%MMB 
& 87.6%MMB-CN
& 84.8%MM-Vet 
& 75.1%MMMU 
& 58.9%MMMU-Pro 
& 71.1%MMStar 
& 61.3%BLINK 
& 79.1%SEED
& 72.5%SEED-2-Plus 
& 69.2%RWQA
& 77.1\\ %Avg
\midrule
InternVL3.5-4B % Method Name
& 82.6%AI2D 
& 86.0%ChartQA 
& 77.1 %MathVista 
& 86.9%MMB 
& 86.1%MMB-CN
& 76.6%MM-Vet 
& 66.6%MMMU 
& 51.4%MMMU-Pro 
& 65.0%MMStar 
& 58.1%BLINK 
& 75.2%SEED
& 69.4%SEED-2-Plus 
& 66.3%RWQA
& 72.9\\ %Avg
\cdashline{1-15}\noalign{\vskip 0.5ex}
$+$Large Teacher% Method Name
& 82.8%AI2D 
& 86.4%ChartQA 
& 77.3%MathVista 
& 86.9%MMB 
& 86.2%MMB-CN
& 76.8%MM-Vet 
& 66.9%MMMU 
& 51.7%MMMU-Pro 
& 65.4%MMStar 
& 58.4%BLINK 
& 75.5%SEED
& 69.7%SEED-2-Plus 
& 66.6%RWQA
& 73.1\\ %Avg
$+$Mid Teacher% Method Name
& 82.9%AI2D 
& 86.6%ChartQA 
& 77.5%MathVista 
& 87.0%MMB 
& 86.3%MMB-CN
& 77.0%MM-Vet 
& 67.1%MMMU 
& 51.9%MMMU-Pro 
& 65.6%MMStar 
& 58.6%BLINK 
& 75.7%SEED
& 69.9%SEED-2-Plus 
& 66.8%RWQA
& 73.3\\ %Avg
$+$Mask-Progressive% Method Name
& 83.1%AI2D 
& 86.9%ChartQA 
& 77.9%MathVista 
& \textbf{87.2}%MMB 
& \textbf{86.5}%MMB-CN
& 77.4%MM-Vet 
& \textbf{67.5}%MMMU 
& 52.2%MMMU-Pro 
& 66.0%MMStar 
& 59.0%BLINK 
& 76.1%SEED
& 70.3%SEED-2-Plus 
& 67.2%RWQA
& 73.6\\ %Avg
\rowcolor{colorful}
$+$Reward Feedback% Method Name
& \textbf{83.3}%AI2D 
& \textbf{94.9}%ChartQA 
& \textbf{78.8}%MathVista 
& 86.1%MMB 
& 86.0%MMB-CN
& \textbf{79.5}%MM-Vet 
& 63.2%MMMU 
& \textbf{52.7}%MMMU-Pro 
& \textbf{70.5}%MMStar 
& \textbf{62.5}%BLINK 
& \textbf{79.1}%SEED
& \textbf{71.4}%SEED-2-Plus 
& \textbf{68.1} %RWQA
& \textbf{75.1}\\ %Avg
\rowcolor{colorful2}
$-$Mid Teacher% Method Name
& 82.9%AI2D 
& 87.0%ChartQA 
& 77.9%MathVista 
& 86.8%MMB 
& 86.3%MMB-CN
& 77.9%MM-Vet 
& 66.9%MMMU 
& 52.2%MMMU-Pro 
& 65.8%MMStar 
& 59.0%BLINK 
& 75.9%SEED
& 70.1%SEED-2-Plus 
& 66.9%RWQA
& 73.5\\ %Avg
\midrule
InternVL3.5-2B % Method Name
& 78.8%AI2D 
& 80.7%ChartQA 
& 71.8 %MathVista 
& 82.3%MMB 
& 81.8%MMB-CN
& 71.7%MM-Vet 
& 59.0%MMMU 
& 46.3%MMMU-Pro 
& 62.7%MMStar 
& 51.3 %BLINK 
& 75.2%SEED
& 68.0%SEED-2-Plus 
& 62.0%RWQA
& 68.6\\ %Avg
\cdashline{1-15}\noalign{\vskip 0.5ex}
$+$Large Teacher% Method Name
& 79.1%AI2D 
& 81.8%ChartQA 
& 72.3%MathVista 
& 82.6%MMB 
& 82.2%MMB-CN
& 72.5%MM-Vet 
& 59.4%MMMU 
& 46.8%MMMU-Pro 
& 63.3%MMStar 
& 52.1%BLINK 
& 75.5%SEED
& 68.3%SEED-2-Plus 
& 62.5%RWQA
& 69.1\\ %Avg
$+$Mid Teacher% Method Name
& 79.7%AI2D 
& 83.8%ChartQA 
& 73.0%MathVista 
& 83.1%MMB 
& 82.8%MMB-CN
& 73.8%MM-Vet 
& 60.2%MMMU 
& 47.6%MMMU-Pro 
& 64.4%MMStar 
& 53.4%BLINK 
& 76.0%SEED
& 68.8%SEED-2-Plus 
& 63.3%RWQA
& 70.0\\ %Avg
$+$Mask-Progressive% Method Name
& 80.9%AI2D 
& 87.7%ChartQA 
& 74.6%MathVista 
& 84.1%MMB 
& 84.0%MMB-CN
& 76.5%MM-Vet 
& 61.8%MMMU 
& 49.3%MMMU-Pro 
& 66.6%MMStar 
& 56.1%BLINK 
& 77.1%SEED
& 69.8%SEED-2-Plus 
& 65.0%RWQA
& 71.8\\ %Avg
\rowcolor{colorful}
$+$Reward Feedback% Method Name
& \textbf{83.0}%AI2D 
& \textbf{94.8}%ChartQA 
& \textbf{77.5}%MathVista 
& \textbf{85.9}%MMB 
& \textbf{86.3}%MMB-CN
& \textbf{81.4}%MM-Vet 
& \textbf{64.6}%MMMU 
& \textbf{52.4}%MMMU-Pro 
& \textbf{70.7}%MMStar 
& \textbf{61.1}%BLINK 
& \textbf{79.0}%SEED
& \textbf{71.6}%SEED-2-Plus 
& \textbf{68.1} %RWQA
& \textbf{75.1}\\ %Avg
\rowcolor{colorful2}
$-$Mid Teacher% Method Name
& 80.1%AI2D 
& 84.3%ChartQA 
& 73.4%MathVista 
& 83.5%MMB 
& 83.2%MMB-CN
& 74.2%MM-Vet 
& 60.6%MMMU 
& 47.9%MMMU-Pro 
& 64.8%MMStar 
& 53.7%BLINK 
& 76.3%SEED
& 69.0%SEED-2-Plus 
& 63.6%RWQA
& 70.4\\ %Avg
% InternVL3.5-1B % Method Name
% & 71.1%AI2D 
% & 77.7%ChartQA 
% & 59.3%MathVista 
% & 74.5%MMB 
% & 74.0%MMB-CN
% & 56.5%MM-Vet 
% & 44.2 %MMMU 
% & 19.7%MMMU-Pro 
% & 51.9%MMStar 
% & 44.0%BLINK 
% & 72.1%SEED
% & 62.3%SEED-2-Plus 
% & 57.6%RWQA
% & \\ %Avg
% +Large Teacher% Method Name
% & %AI2D 
% & %ChartQA 
% & %MathVista 
% & %MMB 
% & %MMB-CN
% & %MM-Vet 
% & %MMMU 
% & %MMMU-Pro 
% & %MMStar 
% & %BLINK 
% & %SEED
% & %SEED-2-Plus 
% & %RWQA
% & \\ %Avg
% +Mid Teacher% Method Name
% & %AI2D 
% & %ChartQA 
% & %MathVista 
% & %MMB 
% & %MMB-CN
% & %MM-Vet 
% & %MMMU 
% & %MMMU-Pro 
% & %MMStar 
% & %BLINK 
% & %SEED
% & %SEED-2-Plus 
% & %RWQA
% & \\ %Avg
% +Master w.o. RL% Method Name
% & %AI2D 
% & %ChartQA 
% & %MathVista 
% & %MMB 
% & %MMB-CN
% & %MM-Vet 
% & %MMMU 
% & %MMMU-Pro 
% & %MMStar 
% & %BLINK 
% & %SEED
% & %SEED-2-Plus 
% & %RWQA
% & \\ %Avg
% \rowcolor{colorful}
% +Reward Feedback% Method Name
% & %AI2D 
% & %ChartQA 
% & %MathVista 
% & %MMB 
% & %MMB-CN
% & %MM-Vet 
% & %MMMU 
% & %MMMU-Pro 
% & %MMStar 
% & %BLINK 
% & %SEED
% & %SEED-2-Plus 
% & %RWQA
% & \\ %Avg
\bottomrule
\end{tabular}
}
\vspace{0mm}
\end{table*}

\section{Masters}
\label{sec:method}

This section introduces the two pivotal components that form the core of the \textbf{Masters}: how to mask the teacher (\cref{sec:mask,sec:progressive}) and reinforce the student (\cref{sec:reinforce}).

\subsection{Magnitude-based Teacher Masking}
\label{sec:mask}
A key challenge in distilling knowledge from a large teacher to a small student lies in the significant parameter gap between them. To reduce this gap, we adopt a magnitude-based masking strategy inspired by classical network pruning~\cite{han2015learning}, where weights with smaller magnitudes are masked to zero. Given a teacher $\mathcal{T}$ with weight $\mathbf{W}_\mathcal{T} = \{w_n\}_{n=1}^{N}$ ($w_n \in \mathbb{R}$), we construct a binary mask $\mathbf{M}_r = \{m_n\}_{n=1}^{N}$ ($m_n \in {0,1}$) as follows:
\begin{equation}
m_n =
\begin{cases}
1, & \text{if } |w_n| \ge \lambda_r, \\
0, & \text{otherwise},
\end{cases}
\end{equation}
where $N$ denotes the total number of parameters (e.g., $N=38$B when using InternVL3.5-38B~\cite{wang2025internvl3}), and $\lambda_r$ is a magnitude threshold determined by the desired masking ratio $r \in [0,1]$. For example, when $r = 0$, it is the original teacher. Conversely, when $r = 1$, all weights are masked to zero. With $r = 0.2$, the magnitude threshold $\lambda_{0.2}$ is determined by sorting the magnitude of $\mathbf{W}_\mathcal{T}$ in ascending order, such that approximately $\sum_{n=1}^N m_n \approx N \cdot (1 - 0.2)$. The resulting masked teacher parameters are defined as:
\begin{equation}
\mathbf{W}_{\mathcal{T}_r} = \mathbf{M}_r \odot \mathbf{W}_\mathcal{T},
\end{equation}
where $\odot$ denotes element-wise multiplication, and $\mathcal{T}_r$ represents the masked teacher under masking ratio $r$. This masking process effectively removes low-magnitude weights that contribute marginally to prediction logits~\cite{han2015learning}, yielding a simplified yet representative teacher. By reducing parameter of the teacher, the student $\mathcal{S}$ learns a more capacity-aligned representation, mitigating optimization instability. Notably, unlike conventional pruning approaches~\cite{lecun1989optimal, hassibi1993optimal, he2017channel, wang2019eigendamage, singh2020woodfisher, lee2022masking, he2023structured, cheng2024survey} designed for model compression, our masking is temporary and restored later. In practice, we found that using a global threshold $\lambda_r$ across all layers often excessively prunes certain layers, which can make the model non-functional at inference time. To prevent this imbalance, we compute $\lambda_r$ per layer and apply masking separately to each layer so that the overall masking remains balanced and consistent across the teacher network.

% table 3
\begin{table*}[t!]
\vspace{0mm}
\centering
\caption{Comparing \textbf{Masters}-applied VLMs with standard or smaller model size open-source VLMs.}
\vspace{-3mm}
\label{tab:3}
\resizebox{\linewidth}{!}{
\renewcommand{\tabcolsep}{1mm}
\begin{tabular}{lccccccccccccc}
\toprule
VLMs
& AI2D
& ChartQA
& MathVista
& MMB
& MMB$^{\text{CN}}$
& MM-Vet
& MMMU
& MMMU-Pro
& MMStar
& BLINK
& SEED
& SEED2+
& RWQA\\
\midrule
LLaVA-OneVision-7B~\cite{li2024llava}
& 81.4 % AI2D
& 80.0% % ChartQA
& 63.2  % MathVista
& 80.8 %MMB
& - %MMB CN
& 57.5 %MM-Vet
& 48.8 %MMMU   
& 24.1%MMMU-Pro
& 61.9%MMStar
& 53.0%BLINK
& 76.7% SEED
& 65.4%Seed-2-plus
& 69.9\\ %RWQA
LLaVA-OneVision-1.5-8B~\cite{an2025llava}
& 84.2% AI2D
& 86.5% ChartQA
& 69.6% MathVista
& 84.1%MMB
& 81.0%MMB CN
& -%MM-Vet
& 55.4 %MMMU   
& 37.4 %MMMU-Pro
& 67.7 %MMStar
& -%BLINK
& 77.3% SEED
& 69.2%Seed-2-plus
& 68.1\\ %RWQA  
MiniCPM-V2.6-8B~\cite{yao2024minicpm}
& 82.1 % AI2D
& -% % ChartQA
& 60.6  % MathVista
& - %MMB
& - %MMB CN
& 60.0 %MM-Vet
& 49.8 %MMMU   
& 27.2%MMMU-Pro
& 57.5%MMStar
& 55.2%BLINK
& 74.0% SEED
& 65.7%Seed-2-plus
& 65.0\\ %RWQA
MiniCPM-o2.6-8B~\cite{yao2024minicpm}
& 86.1% AI2D
& 86.9% % ChartQA
& 73.3% MathVista
& -%MMB
& -%MMB CN
& 67.2%MM-Vet
& 50.9%MMMU   
& -%MMMU-Pro
& 63.3%MMStar
& 53.9%BLINK
& 75.5% SEED
& 67.9%Seed-2-plus
& 68.0\\ %RWQA
Ovis2-8B~\cite{lu2024ovis}
& 86.6% AI2D
& -% ChartQA
& 71.8% MathVista
& -%MMB
& -%MMB CN
& 65.1%MM-Vet
& 57.4%MMMU   
& -%MMMU-Pro
& 64.6%MMStar
& 54.3%BLINK
& 77.2% SEED
& 70.4%Seed-2-plus
& 72.5\\ %RWQA
GLM-4.1V-9B~\cite{hong2025glm}
& 87.9% AI2D
& 70.0% ChartQA
& 80.7% MathVista
& -%MMB
& -%MMB CN
& 66.4%MM-Vet
& 68.0 %MMMU   
& -%MMMU-Pro
& 72.9%MMStar
& 65.9%BLINK
& -% SEED
& 71.8%Seed-2-plus
& 70.6\\ %RWQA
MiMo-VL-8B~\cite{coreteam2025mimovltechnicalreport}
& 83.5% AI2D
& 91.7% ChartQA
& 81.5% MathVista
& 84.4%MMB
& 82.0%MMB CN
& 77.5%MM-Vet
& 66.7%MMMU   
& 46.2%MMMU-Pro
& 70.8%MMStar
& 62.4%BLINK
& -% SEED
& 72.4%Seed-2-plus
& -\\ %RWQA
Keye-VL-8B~\cite{team2025kwai}
& 86.7% AI2D
& 72.5% ChartQA
& 80.7% MathVista
& \textbf{92.0}%MMB
& -%MMB CN
& 68.6%MM-Vet
& 71.4%MMMU   
& -%MMMU-Pro
& 75.5%MMStar
& 52.0%BLINK
& -% SEED
& 67.8%Seed-2-plus
& 67.7\\ %RWQA
Keye-VL-1.5-8B~\cite{yang2025kwai}
& \textbf{89.5}% AI2D
& 86.3% ChartQA
& 81.2% MathVista
& \textbf{92.0}%MMB
& -%MMB CN
& 71.2%MM-Vet
& 71.4%MMMU   
& -%MMMU-Pro
& 80.5 %MMStar
& 54.9%BLINK
& -% SEED
& -%Seed-2-plus
& 73.5\\ %RWQA 
Qwen2.5-VL-7B~\cite{bai2025qwen2}
& 83.9%AI2D 
& 87.3%ChartQA 
& 67.8%MathVista 
& 83.5%MMB 
& 83.4%MMB-CN
& 71.8%MM-Vet 
& 55.0%MMMU 
& 38.3%MMMU-Pro 
& 63.9%MMStar 
& 56.4%BLINK 
& 77.0%SEED
& 70.4%SEED-2-Plus 
& 68.5\\ %RWQA
Qwen3-VL-8B~\cite{yang2025qwen3} % Method Name
& 85.7%AI2D 
& 88.4%ChartQA 
& 77.2%MathVista 
& 86.8%MMB 
& 86.5%MMB-CN
& 74.5%MM-Vet 
& 69.6%MMMU 
& 55.9%MMMU-Pro 
& 70.9%MMStar 
& 69.1%BLINK 
& 78.5%SEED
& 70.8%SEED-2-Plus 
& 69.9\\%RWQA
InternVL3-8B~\cite{zhu2025internvl3}
& 85.2%AI2D 
& 86.6%ChartQA 
& 71.6%MathVista 
& 83.4%MMB 
& 82.2%MMB-CN
& 78.5%MM-Vet 
& 62.7%MMMU 
& 41.3%MMMU-Pro 
& 68.2%MMStar 
& 55.5%BLINK 
& 77.1%SEED
& 69.7%SEED-2-Plus 
& 70.8\\ %RWQA
InternVL3.5-8B~\cite{wang2025internvl3} % Method Name
& 84.0%AI2D 
& 86.7%ChartQA 
& 78.4%MathVista 
& 86.5%MMB 
& 85.9%MMB-CN
& 83.1%MM-Vet 
& 73.4%MMMU 
& 57.2%MMMU-Pro 
& 69.3 %MMStar 
& 59.5 %BLINK 
& 77.4%SEED
& 70.8%SEED-2-Plus 
& 67.5\\%RWQA
\midrule
\rowcolor{colorful}
Qwen2.5-VL-7B-\textbf{Masters}
& 88.6%AI2D 
& 95.6%ChartQA 
& 78.8%MathVista 
& 89.1%MMB 
& 89.5%MMB-CN
& 81.7%MM-Vet 
& 71.3%MMMU 
& 60.6%MMMU-Pro 
& 74.9%MMStar 
& 67.2%BLINK 
& 81.8%SEED
& \textbf{75.9}%SEED-2-Plus 
& 77.3\\ %RWQA
\rowcolor{colorful}
Qwen3-VL-8B-\textbf{Masters}
& 88.5%AI2D 
& \textbf{95.9}%ChartQA 
& 81.8%MathVista 
& 89.5%MMB 
& 89.9%MMB-CN
& 79.4%MM-Vet 
& 72.9%MMMU 
& \textbf{61.4}%MMMU-Pro 
& 79.7%MMStar 
& \textbf{72.3}%BLINK 
& 81.7%SEED
& 75.1%SEED-2-Plus 
& \textbf{77.5}\\ %RWQA
\rowcolor{colorful}
InternVL3-8B-\textbf{Masters}
& 88.9%AI2D 
& 94.8%ChartQA 
& 82.3%MathVista 
& 90.1%MMB 
& \textbf{91.0}%MMB-CN
& 83.8%MM-Vet 
& \textbf{74.0}%MMMU 
& 58.8%MMMU-Pro 
& \textbf{82.0}%MMStar 
& 68.0%BLINK 
& \textbf{82.6}%SEED
& 75.0%SEED-2-Plus 
& 74.8\\ %RWQA
\rowcolor{colorful}
InternVL3.5-8B-\textbf{Masters}
& 87.2%AI2D 
& 95.1%ChartQA 
& \textbf{85.0}%MathVista 
& 88.2%MMB 
& 88.3%MMB-CN
& \textbf{85.6}%MM-Vet 
& 72.7%MMMU 
& 58.1%MMMU-Pro 
& 80.8%MMStar 
& 67.8%BLINK 
& 81.4%SEED
& 75.5%SEED-2-Plus 
& 74.9\\ %RWQA
\midrule
% VILA1.5-3B~\cite{lin2023vila}
% & 57.9 % AI2D
% & - % ChartQA
% & 31.6 % MathVista
% & - %MMB
% & - %MMB CN
% & 38.8 %MM-Vet
% & 34.2 %MMMU
% & - %MMMU-Pro
% & 40.6%MMStar
% & 39.7% BLINK
% & 68.0% SEED
% & 41.4 % SEED-2-Plus
% & 53.2 \\ % RWQA 
Phi-3.5-Vision-4B~\cite{abdin2024phi}
& 77.8% AI2D
& 81.8 % ChartQA
& 43.9 % MathVista
& 76.0 %MMB
& 66.1 %MMB CN
& 43.2 %MM-Vet
& 43.0 %MMMU
& 19.7 %MMMU-Pro
& 47.5%MMStar
& 58.3% BLINK
& 69.7% SEED
& 62.2% SEED-2-Plus
& 53.6\\ % RWQA
Phi-4-Multimodal-5.6B~\cite{abouelenin2025phi}
& 83.0 % AI2D
& 81.4 % ChartQA
& 65.8 % MathVista
& 86.7 %MMB
& - %MMB CN
& 51.9 %MM-Vet
& 56.0 %MMMU
& 38.5 %MMMU-Pro
& 58.9%MMStar
& 42.1% BLINK
& 73.2% SEED
& 68.5% SEED-2-Plus
& 64.1\\ % RWQA
LLaVA-OneVision-1.5-4B~\cite{an2025llava}
& 83.6 % AI2D
& 87.1% ChartQA
& 67.9% MathVista
& 84.2%MMB
& 76.9%MMB CN
& -%MM-Vet
& 52.7 %MMMU   
& 35.3 %MMMU-Pro
& 64.9 %MMStar
& -%BLINK
& 76.6% SEED
& 68.9%Seed-2-plus
& 67.8\\ %RWQA  
Ovis2-4B~\cite{lu2024ovis}
& 85.7% AI2D
& -% ChartQA
& 69.6% MathVista
& -%MMB
& -%MMB CN
& 65.5%MM-Vet
& 49.0%MMMU   
& -%MMMU-Pro
& 61.9%MMStar
& 53.0%BLINK
& 76.2% SEED
& 69.3%Seed-2-plus
& 71.1\\ %RWQA 
Qwen2.5-VL-3B~\cite{bai2025qwen2}
& 81.6%AI2D 
& 84.0%ChartQA 
& 61.2%MathVista 
& 79.1%MMB 
& 78.1%MMB-CN
& 61.8%MM-Vet 
& 51.2%MMMU 
& 31.6%MMMU-Pro 
& 55.9%MMStar 
& 47.6%BLINK 
& 74.0%SEED
& 67.6%SEED-2-Plus 
& 65.4\\%RWQA
Qwen3-VL-4B~\cite{yang2025qwen3} % Method Name
& 84.1%AI2D 
& 87.4%ChartQA 
& 73.7%MathVista 
& 84.6%MMB 
& 84.3%MMB-CN
& 71.0%MM-Vet 
& 67.4%MMMU 
& 53.2%MMMU-Pro 
& 69.8%MMStar 
& 65.8%BLINK 
& 78.4%SEED
& 70.2%SEED-2-Plus 
& 70.6 \\%RWQA
InternVL3.5-4B~\cite{wang2025internvl3} % Method Name
& 82.6%AI2D 
& 86.0%ChartQA 
& 77.1 %MathVista 
& 86.9%MMB 
& 86.1%MMB-CN
& 76.6%MM-Vet 
& 66.6%MMMU 
& 51.4%MMMU-Pro 
& 65.0%MMStar 
& 58.1%BLINK 
& 75.2%SEED
& 69.4%SEED-2-Plus 
& 66.3\\%RWQA
\midrule
\rowcolor{colorful}
Qwen2.5-VL-3B-\textbf{Masters}
& 85.4%AI2D 
& \textbf{95.6}%ChartQA 
& 75.6%MathVista 
& 85.9%MMB 
& 86.3%MMB-CN
& 78.3%MM-Vet 
& 61.2%MMMU 
& 51.0%MMMU-Pro 
& 68.9%MMStar 
& 53.0%BLINK 
& 78.8%SEED
& 72.1%SEED-2-Plus 
& 70.7\\ %RWQA
\rowcolor{colorful}
Qwen3-VL-4B-\textbf{Masters}
& \textbf{88.0}%AI2D 
& 94.7%ChartQA 
& \textbf{79.6}%MathVista 
& \textbf{88.7}%MMB 
& \textbf{89.3}%MMB-CN
& 79.4%MM-Vet 
& \textbf{70.3}%MMMU 
& \textbf{57.0}%MMMU-Pro 
& \textbf{75.3}%MMStar 
& \textbf{69.1}%BLINK 
& \textbf{81.5}%SEED
& \textbf{72.9}%SEED-2-Plus 
& \textbf{77.8}\\ %RWQA
\rowcolor{colorful}
InternVL3.5-4B-\textbf{Masters}
& 83.3%AI2D 
& 94.9%ChartQA 
& 78.8%MathVista 
& 86.1%MMB 
& 86.0%MMB-CN
& \textbf{79.5}%MM-Vet 
& 63.2%MMMU 
& 52.7%MMMU-Pro 
& 70.5%MMStar 
& 62.5%BLINK 
& 79.1%SEED
& 71.4%SEED-2-Plus 
& 68.1\\ %RWQA
\midrule
Aquila-VL-2B~\cite{chen2024expanding}
& 75.0% AI2D
& 76.5% ChartQA
& 59.0% MathVista
& -%MMB
& -%MMB CN
& 43.8%MM-Vet
& 47.4%MMMU
& -%MMMU-Pro
& 54.9%MMStar
& 34.1% BLINK
& 73.9% SEED
& 63.0% SEED-2-Plus
& 65.0\\ % RWQA
Ovis2-2B~\cite{lu2024ovis}
& 82.7% AI2D
& -% ChartQA
& 64.1% MathVista
& -%MMB
& -%MMB CN
& 58.3%MM-Vet
& 45.6%MMMU   
& -%MMMU-Pro
& 56.7%MMStar
& 47.9%BLINK
& 74.4% SEED
& 67.4%Seed-2-plus
& 66.0\\ %RWQA 
Qwen3-VL-2B~\cite{yang2025qwen3}
& 76.9%AI2D 
& 81.2%ChartQA 
& 61.3%MathVista 
& 81.9%MMB 
& 81.4%MMB-CN
& 61.4%MM-Vet 
& 53.4%MMMU 
& 36.5%MMMU-Pro 
& 58.3%MMStar 
& 53.8%BLINK 
& 76.0%SEED
& 66.4%SEED-2-Plus 
& 63.9\\ %RWQA
InternVL3-2B~\cite{zhu2025internvl3}
& 78.7%AI2D 
& 80.2%ChartQA 
& 57.0%MathVista 
& 81.1%MMB 
& 78.4%MMB-CN
& 62.2%MM-Vet 
& 48.6%MMMU 
& 24.9%MMMU-Pro 
& 60.7%MMStar 
& 47.0%BLINK 
& 75.0%SEED
& 64.6%SEED-2-Plus 
& 64.3\\ %RWQA
InternVL3.5-2B~\cite{wang2025internvl3} % Method Name
& 78.8%AI2D 
& 80.7%ChartQA 
& 71.8 %MathVista 
& 82.3%MMB 
& 81.8%MMB-CN
& 71.7%MM-Vet 
& 59.0%MMMU 
& 46.3%MMMU-Pro 
& 62.7%MMStar 
& 51.3 %BLINK 
& 75.2%SEED
& 68.0%SEED-2-Plus 
& 62.0\\%RWQA
\midrule
\rowcolor{colorful}
Qwen3-VL-2B-\textbf{Masters}
& 82.1%AI2D 
& 94.1%ChartQA 
& 70.4%MathVista 
& 84.2%MMB 
& 84.1%MMB-CN
& 71.3%MM-Vet 
& 55.8%MMMU 
& 39.5%MMMU-Pro 
& 66.1%MMStar 
& \textbf{62.3}%BLINK 
& 78.3%SEED
& 69.4%SEED-2-Plus 
& 69.0\\ %RWQA
\rowcolor{colorful}
InternVL3-2B-\textbf{Masters}
& 82.1%AI2D 
& 93.5%ChartQA 
& 66.5%MathVista 
& \textbf{85.9}%MMB 
& 85.2%MMB-CN
& 77.0%MM-Vet 
& 59.6%MMMU 
& 40.1%MMMU-Pro 
& 66.7%MMStar 
& 53.1%BLINK
& 78.5%SEED
& 67.8%SEED-2-Plus 
& \textbf{68.4}\\ %RWQA
\rowcolor{colorful}
InternVL3.5-2B-\textbf{Masters}
& \textbf{83.0}%AI2D 
& \textbf{94.8}%ChartQA 
& \textbf{77.5}%MathVista 
& \textbf{85.9}%MMB 
& \textbf{86.3}%MMB-CN
& \textbf{81.4}%MM-Vet 
& \textbf{64.6}%MMMU 
& \textbf{52.4}%MMMU-Pro 
& \textbf{70.7}%MMStar 
& 61.1%BLINK 
& \textbf{79.0}%SEED
& \textbf{71.6}%SEED-2-Plus 
& 68.1\\ %RWQA
% \midrule
% LLaVA-OneVision-0.5B~\cite{li2024llava}
% & 57.1% AI2D  
% & 61.4% ChQA
% & 34.8% MatV
% & 61.6% MMB
% & 55.5% MMBC
% & 32.2% MMVet
% & 31.4%MMMU
% & -%MMMU-Pro
% & 37.7%MMStar
% & 52.1% BLINK
% & 66.6% SEED
% & 45.7% SEED-2-Plus
% & 55.6\\ % RWQA
% Ovis2-1B~\cite{lu2024ovis}
% & 76.4% AI2D
% & -% ChartQA
% & 59.4% MathVista
% & -%MMB
% & -%MMB CN
% & 50.0%MM-Vet
% & 36.1%MMMU   
% & -%MMMU-Pro
% & 52.1%MMStar
% & 44.0%BLINK
% & 71.7% SEED
% & 61.4%Seed-2-plus
% & 63.9\\ %RWQA 
% InternVL3-1B~\cite{zhu2025internvl3}
% & 69.4%AI2D 
% & 75.3%ChartQA 
% & 45.8%MathVista 
% & 72.6%MMB 
% & 67.9%MMB-CN
% & 58.7%MM-Vet 
% & 43.4%MMMU 
% & 17.5%MMMU-Pro 
% & 51.5%MMStar 
% & 42.9%BLINK 
% & 71.2%SEED
% & 58.2%SEED-2-Plus 
% & 58.2\\ %RWQA
% InternVL3.5-1B~\cite{wang2025internvl3} % Method Name
% & 71.1%AI2D 
% & 77.7%ChartQA 
% & 59.3%MathVista 
% & 74.5%MMB 
% & 74.0%MMB-CN
% & 56.5%MM-Vet 
% & 44.2 %MMMU 
% & 19.7%MMMU-Pro 
% & 51.9%MMStar 
% & 44.0%BLINK 
% & 72.1%SEED
% & 62.3%SEED-2-Plus 
% & 57.6\\%RWQA
% \midrule
% \rowcolor{colorful}
% InternVL3-\textbf{Masters}-1B
% & %AI2D 
% & %ChartQA 
% & %MathVista 
% & %MMB 
% & %MMB-CN
% & %MM-Vet 
% & %MMMU 
% & %MMMU-Pro 
% & %MMStar 
% & %BLINK 
% & %SEED
% & %SEED-2-Plus 
% & \\ %RWQA
% \rowcolor{colorful}
% InternVL3.5-\textbf{Masters}-1B
% & %AI2D 
% & %ChartQA 
% & %MathVista 
% & %MMB 
% & %MMB-CN
% & %MM-Vet 
% & %MMMU 
% & %MMMU-Pro 
% & %MMStar 
% & %BLINK 
% & %SEED
% & %SEED-2-Plus 
% & \\ %RWQA
\bottomrule
\end{tabular}
}
\vspace{0mm}
\end{table*}

% figure 4
\begin{figure}[t!]
\vspace{0mm}
    \centering
    \includegraphics[width=0.35\textwidth]{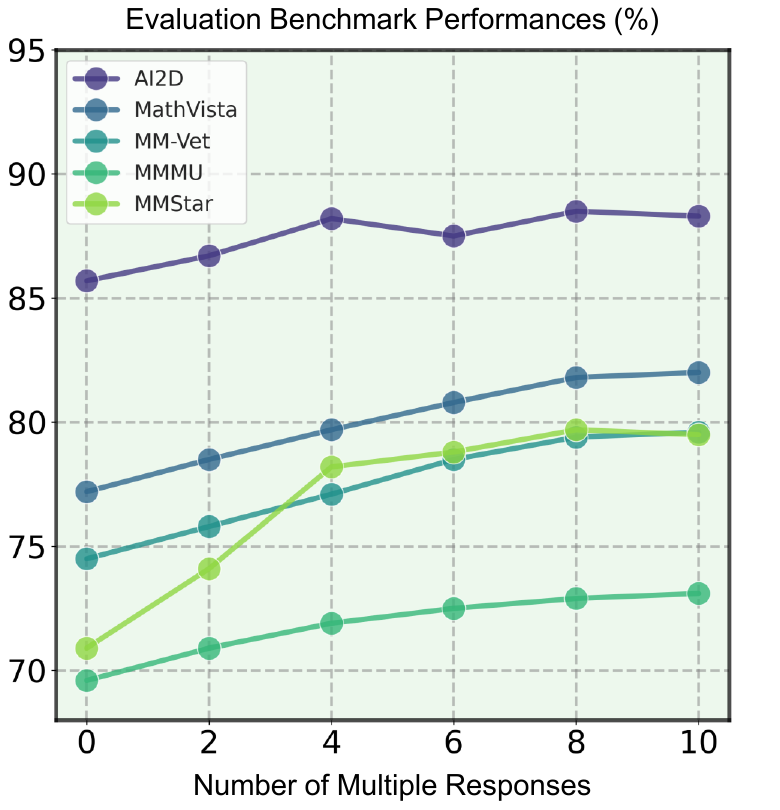}
    \vspace{0mm}
    \caption{Comparing the performances by the number of generated responses from teacher and student.}
    \label{fig:4}
    \vspace{0mm}
\end{figure}

\subsection{Mask-Progressive Distillation}
\label{sec:progressive}
\vspace{1.5mm}
\noindent\textbf{Scheduling Masking Ratio.}
While the masked teacher provides a simplified representation, learning solely with a fixed masking ratio limits the student’s exposure to the teacher’s rich representation. To solve it, we adopt a \textit{mask-progressive} strategy that gradually restores the teacher’s full capacity over the entire training. At each training iteration $i \in \{1, \ldots, I\}$, a masking ratio $r[i]$ is formulated as written:
\begin{equation}
r[i] = r_{\max} - s \cdot \left\lfloor i \times \frac{M}{I}\right\rfloor,
\label{eq:3}
\end{equation}
where $s$ denotes the decrement applied to the masking ratio at each masking stage, and $M$ represents the total number of masked teachers with different masking ratios, computed as $M = r_{\max} / s + 1$. For example, setting $r_{\max} = 0.2$ and $s = 0.05$ results in $M = 5$, meaning that the masking ratio progressively decreases as $0.20$, $0.15$, $0.10$, $0.05$, and finally $0$. \cref{fig:2}(a) illustrates how the masking ratio $r[i]$ is scheduled during training.

As the masking ratio changes, the distillation objective also adapt to reflect the teacher’s gradually increasing capacity. Formally, the distillation objective at iteration $i$ is formulated as:
\begin{equation}
\min_{\mathbf{W}_\mathcal{S}}\mathbb{E}_{(x,y)\sim \text{Data}[i]}\left[\mathcal{D}\left( 
P_{\mathcal{T}_{r[i]}}(y|x) \, \Vert \, P_{\mathcal{S}}(y|x)\right)
\right],
\label{eq:4}
\end{equation}
where $\mathbf{W}_\mathcal{S}$ indicates the weight set of the student $\mathcal{S}$, and $P(y|x)$ denotes the logit-softmax output for answer label $y$ given a question $x$ under the data subset $\text{Data}[i]$ at the iteration $i$. Note that $\mathcal{D}$ refers to Jensen-Shannon Divergence (JSD)~\cite{xu2024llavadi, agarwal2024policy}, which has been empirically shown to outperform the standard KL divergence for distillation tasks. This mask-progressive distillation allows the student to first learn coarse-grained knowledge and progressively refine its representations as the teacher’s representational capacity is restored throughout training.

\vspace{1.5mm}
\noindent\textbf{Multiple Responses.}
While mask-progressive distillation enables the student to learn from a capacity-aligned teacher, the training data samples used for distillation still exhibit two inherent limitations. First, conventional SFT datasets are typically generated by large closed-source models~\cite{hurst2024gpt, comanici2025gemini} or filtered by humans~\cite{xu2024vision}. Injecting such responses into small student often leads to performance degradation due to representation mismatch—small vocabulary and low hidden dimension of representation. Second, standard SFT datasets generally provide only a single answer per question, which severely limits response diversity and generalization. This single-answer guidance forces the student to overfit to a narrow linguistic or reasoning style, reducing its versatility across broader vision–language contexts.

% table 5
\begin{table*}[h!]
\vspace{0mm}
\centering
\caption{Comparing Masters-applied Small VLMs with Closed-source and Large Open-source VLMs.}
\label{tab:5}
\vspace{-3mm}
\resizebox{\linewidth}{!}{
\renewcommand{\tabcolsep}{2mm}
\begin{tabular}{lccccccccccccc}
\toprule
VLMs
& AI2D 
& ChartQA
& MathVista 
& MMB 
& MM-Vet 
& MMMU 
& MMMU-Pro 
& MMStar 
& BLINK 
& SEED
& SEED2+
& RWQA \\
\midrule
Claude-3.5-Sonnet~\cite{claude3series2024}
& 81.2%AI2D 
& 90.8%ChartQA 
& 67.7%MathVista 
& 82.6%MMB 
& 70.1%MM-Vet 
& 68.3%MMMU 
& 51.5 %MMMU-Pro 
& 65.1%MMStar 
& 60.1%BLINK 
& 61.7%SEED
& 71.7%SEED-2-Plus 
& 65.8\\ %RWQA
Claude-3.7-Sonnet~\cite{claude3series2024}
& 82.5%AI2D 
& 92.2%ChartQA 
& 66.8%MathVista 
& 84.8%MMB 
& 70.0%MM-Vet 
& 71.0%MMMU 
& 56.5%MMMU-Pro 
& 65.1%MMStar 
& 56.6%BLINK 
& 74.3%SEED
& 67.6%SEED-2-Plus 
& 55.4\\ %RWQA
Claude-4-Sonnet~\cite{claude3series2024}
& 83.0%AI2D 
& -%ChartQA 
& 74.6%MathVista 
& -%MMB 
& -%MM-Vet 
& 74.4%MMMU 
& 61.6%MMMU-Pro 
& 69.4%MMStar 
& 60.4%BLINK 
& -%SEED
& -%SEED-2-Plus 
& 69.8\\ %RWQA
Gemini-1.5-Pro~\cite{team2023gemini}
& 79.1%AI2D 
& 87.2%ChartQA 
& 63.9%MathVista 
& 73.9%MMB 
& 64.0%MM-Vet 
& 62.2%MMMU 
& 46.9%MMMU-Pro 
& 59.1 %MMStar 
& 59.1%BLINK 
& 76.0%SEED
& 70.8%SEED-2-Plus 
& 67.5\\ %RWQA
Gemini-2.0-Flash~\cite{team2023gemini}
& 83.1%AI2D 
& -%ChartQA 
& 70.4%MathVista 
& 90.0%MMB 
& 73.6%MM-Vet 
& 69.9%MMMU 
& 54.4%MMMU-Pro 
& 69.4%MMStar 
& 64.0%BLINK 
& -%SEED
& -%SEED-2-Plus 
& 72.3\\ %RWQA
Gemini-2.5-Pro~\cite{comanici2025gemini}
& 89.5% AI2D
& -% ChartQA
& 80.9% MathVista
& -%MMB
& 83.3%MM-Vet
& 74.7%MMMU   
& -%MMMU-Pro
& 73.6%MMStar
& -%BLINK
& -% SEED
& -%Seed-2-plus
& -\\ %RWQA
% Gemini-2.5-Pro~\cite{comanici2025gemini}
% & % AI2D
% & % ChartQA
% & % MathVista
% & %MMB
% & %MM-Vet
% & %MMMU   
% & %MMMU-Pro
% & %MMStar
% & %BLINK
% & % SEED
% & %Seed-2-plus
% & \\ %RWQA
GLM-4.5V~\cite{hong2025glm}
& 88.1% AI2D
& 86.6% ChartQA
& 84.6 % MathVista
& -%MMB
& 75.2%MM-Vet
& 75.4 %MMMU   
& 65.2 %MMMU-Pro
& 75.3%MMStar
& 65.3 %BLINK
& -% SEED
& 74.0%Seed-2-plus
& \\ %RWQA
GPT-4o~\cite{gptsyscard}
& 84.6%AI2D 
& 85.7%ChartQA 
& 63.8%MathVista 
& 83.4%MMB 
& 69.1%MM-Vet 
& 69.1%MMMU 
& 51.9%MMMU-Pro 
& 64.7%MMStar 
& 68.0%BLINK 
& 77.1%SEED
& 72.0%SEED-2-Plus 
& 75.4\\ %RWQA
GPT-4.1~\cite{gptsyscard}
& 85.9% AI2D
& -% ChartQA
& 70.4% MathVista
& -%MMB
& 78.8%MM-Vet
& 74.0%MMMU   
& -%MMMU-Pro
& 69.8%MMStar
& \textbf{68.5}%BLINK
& 78.0% SEED
& 73.1%Seed-2-plus
& 78.7\\ %RWQA
GPT-5-Mini~\cite{gptsyscard}
& 88.2% AI2D
& -% ChartQA
& 79.1% MathVista
& -%MMB
& -%MM-Vet
& 79.0%MMMU   
& \textbf{67.3}%MMMU-Pro
& 74.1%MMStar
& -%BLINK
& -% SEED
& -%Seed-2-plus
& \textbf{79.0}\\ %RWQA
GPT-5~\cite{gptsyscard}
& 89.5% AI2D
& -% ChartQA
& 81.9% MathVista
& -%MMB
& 77.6%MM-Vet
& \textbf{84.2}%MMMU   
& -%MMMU-Pro
& 75.7%MMStar
& -%BLINK
& -% SEED
& -%Seed-2-plus
& -\\ %RWQA
\midrule
NVLM-72B~\cite{nvlm2024}
& 85.2%AI2D 
& 86.0%ChartQA 
& 66.6%MathVista 
& -%MMB 
& 58.9%MM-Vet 
& 59.7%MMMU 
& -%MMMU-Pro 
& 63.7 %MMStar 
& 48.0%BLINK 
& 75.5%SEED
& 68.4%SEED-2-Plus 
& 69.9\\ %RWQA
LLaVA-OneVision-72B~\cite{li2024llava}
& 85.6%AI2D 
& 83.7%ChartQA 
& 67.5%MathVista 
& 85.8%MMB 
& 60.6%MM-Vet 
& 56.8%MMMU 
& 31.0%MMMU-Pro 
& 65.8%MMStar 
& 55.4%BLINK 
& 77.5%SEED
& -%SEED-2-Plus 
& 71.9\\ %RWQA
Molmo-72B~\cite{deitke2024molmo}
& 83.4%AI2D 
& 87.3%ChartQA 
& 58.6%MathVista 
& -%MMB 
& 61.1%MM-Vet 
& 54.1%MMMU 
& -%MMMU-Pro 
& 63.3%MMStar 
& 49.7%BLINK 
& 74.6%SEED
& 67.6%SEED-2-Plus 
& 73.7\\ %RWQA
Qwen2.5-VL-72B~\cite{bai2025qwen2}
& 88.7% AI2D
& 89.5% ChartQA
& 74.2% MathVista
& 88.6%MMB
& 76.9%MM-Vet
& 68.2%MMMU   
& 61.2%MMMU-Pro
& 70.8%MMStar
& 64.4%BLINK
& 79.5% SEED
& 73.0%Seed-2-plus
& 75.7\\ %RWQA
Qwen3-VL-32B~\cite{yang2025qwen3}
& 89.5% AI2D
& 89.8% ChartQA
& 83.8% MathVista
& \textbf{90.6}%MMB
& 79.4%MM-Vet
& 76.0%MMMU   
& 65.3%MMMU-Pro
& 77.7%MMStar
& 67.3%BLINK
& 79.9% SEED
& 72.8%Seed-2-plus
& \textbf{79.0}\\ %RWQA
InternVL3-78B~\cite{zhu2025internvl3}
& \textbf{89.7}% AI2D
& 89.7% ChartQA
& 79.0% MathVista
& 89.0%MMB
& 81.3%MM-Vet
& 72.2%MMMU   
& 62.3%MMMU-Pro
& 72.5%MMStar
& 66.3%BLINK
& 78.7% SEED
& 71.9%Seed-2-plus
& 78.0\\ %RWQA
InternVL3.5-38B~\cite{wang2025internvl3}
& 87.8% AI2D
& 88.8% ChartQA
& 81.9 % MathVista
& 90.3 %MMB
& 82.2%MM-Vet
& 76.9%MMMU   
& 66.0%MMMU-Pro
& 75.3%MMStar
& 60.9 %BLINK
& 79.1% SEED
& 71.0%Seed-2-plus
& 75.9\\ %RWQA
\midrule
\rowcolor{colorful}
Qwen2.5-VL-7B-\textbf{Masters}
& 88.6%AI2D 
& 95.6%ChartQA 
& 78.8%MathVista 
& 89.1%MMB 
& 81.7%MM-Vet 
& 71.3%MMMU 
& 60.6%MMMU-Pro 
& 74.9%MMStar 
& 67.2%BLINK 
& 81.8%SEED
& \textbf{75.9}%SEED-2-Plus 
& 77.3\\ %RWQA
\rowcolor{colorful}
Qwen3-VL-8B-\textbf{Masters}
& 88.5%AI2D 
& \textbf{95.9}%ChartQA 
& 81.8%MathVista 
& 89.5%MMB 
& 79.4%MM-Vet 
& 72.9%MMMU 
& 61.4%MMMU-Pro 
& 79.7%MMStar 
& 72.3%BLINK 
& 81.7%SEED
& 75.1%SEED-2-Plus 
& 77.5\\ %RWQA
\rowcolor{colorful}
InternVL3-8B-\textbf{Masters}
& 88.9%AI2D 
& 94.8%ChartQA 
& 82.3%MathVista 
& 90.1%MMB 
& 83.8%MM-Vet 
& 74.0%MMMU 
& 58.8%MMMU-Pro 
& 82.0%MMStar 
& 68.0%BLINK 
& \textbf{82.6}%SEED
& 75.0%SEED-2-Plus 
& 74.8\\ %RWQA
\rowcolor{colorful}
InternVL3.5-8B-\textbf{Masters}
& 87.2%AI2D 
& 95.1%ChartQA 
& \textbf{85.0}%MathVista 
& 88.2%MMB 
& \textbf{85.6}%MM-Vet 
& 72.7%MMMU 
& 58.1%MMMU-Pro 
& \textbf{80.8}%MMStar 
& 67.8%BLINK 
& 81.4%SEED
& 75.5%SEED-2-Plus 
& 74.9\\ %RWQA
\bottomrule
\end{tabular}
}
\vspace{0mm}
\end{table*}

To address these limitations, \textbf{Masters} generates multiple responses from the masked teacher, where the teacher’s representational capacity increases as the masking ratio decreases. These responses provide the student with adaptive guidance signals that evolve in step with their learning capacity. Furthermore, \textbf{Masters} incorporates the student’s own responses into training to maintain alignment between teacher guidance and the student’s evolving representational capacity. The formulation regarding multiple responses is written as follows:
\begin{equation}
\min_{\mathbf{W}_\mathcal{S}}\mathbb{E}_{(x,\hat{y})\sim \text{Gen-Data}[i]}\left[\mathcal{D}\left( 
P_{\mathcal{T}_{r[i]}}(\hat{y}|x) \, \Vert \, P_{\mathcal{S}}(\hat{y}|x)\right)\right],
\label{eq:5}
\end{equation}
where Gen-Data$[i]$ denotes the \textit{pre-generated} multi-response dataset obtained from both the masked teacher and the student at training iteration $i$. This enables stable yet continual improvement toward teacher-level behavior—effectively overcoming the single-answer and over-rich label constraints of conventional SFT datasets.

\subsection{Reinforcing Student with Dual Rewards}
\label{sec:reinforce}
Although generating multiple responses helps mitigate the constraints of conventional SFT datasets, some generated responses may include factual inaccuracies or overly complex language that hampers effective knowledge transfer, thereby reducing distillation performance. To address this, we evaluate both the factual accuracy and the transferability of the responses and refine the student based on their feedback to avoid generating such undesirable responses.

\vspace{1.5mm}
\noindent\textbf{Offline RL Setup.} Conventional online RL paradigms, such as the ``think-answer'' process~\cite{guo2025deepseek, shao2024deepseekmath}, require generating multiple and long responses at each training iteration, incurring substantial computational overhead. To overcome this limitation, \textbf{Masters} adopts an offline RL setup where both the teacher and the student \textit{pre-generate} multiple responses for all questions in the training dataset. This design dramatically reduces computational cost while enabling scaled-up training with diverse responses.

\vspace{1.5mm}
\noindent\textbf{Accuracy Reward.}
To assess the correctness of generated responses, we adopt an LLM-as-a-Judge~\cite{zheng2023judging}, which measures semantic fidelity between the generated responses $\hat{y}$ and the answer labels $y$. The accuracy reward is defined as:
\begin{equation}
\mathcal{R}_{\text{acc}} = \text{LLM-as-a-Judge}(x, \hat{y}, y) \in [0,1].
\end{equation}
This reward guides the student to generate semantically accurate and faithful outputs, regardless of linguistic or stylistic variations. Unlike prior approaches such as DeepSeek-R1~\cite{guo2025deepseek} and its follow-up variants~\cite{yu2025perception, liu2025visual, shen2025vlm, huang2025vision}, which depend on traditional parsers for limited domain evaluation metrics (e.g., math reasoning or spatial grounding), LLM-as-a-Judge method eliminates such constraints. For example, traditional parsers fail when handling open-ended visual questions like ``Describe the person’s emotion in this image''. Even, when comparing ``about five minutes'' (predicted) with ``5'' (answer label), traditional parsing says that the predicted response is wrong answer. On the other hands, LLM-as-a-Judge~\cite{zheng2023judging} delivers consistent and robust judgments across diverse visual question answering (VQA) scenarios — including recognition, OCR, reasoning, chart and document understanding — providing a more generalizable and semantically grounded measure of correctness.

\vspace{1.5mm}
\noindent\textbf{Distillation Reward.}
While the accuracy reward ensures response correctness, it does not capture how effectively the student learns responses. To address this, we introduce a distillation reward that measures logit-level similarity between the teacher and student. Specifically, we compute this reward using $\mathcal{D}$ defined in \cref{eq:4}. Practically, we have observed that the values of $\mathcal{D}$ exhibit small variance across the generated responses due to its low scaling. To stabilize optimization and enhance sensitivity~\cite{schaul2021return}, we apply a reverse min-max normalization~\cite{yuan2024rlexplore} as follows:
\begin{equation}
\mathcal{R}_{\text{distill}} = \frac{\mathcal{D}_{\max} - \mathcal{D}}{\mathcal{D}_{\max} - \mathcal{D}_{\min}},
\end{equation}
where $\mathcal{D}_{\min}$ and $\mathcal{D}_{\max}$ denote the minimum and maximum divergence among the generated responses $\hat{y}$ for a given question $x$. This normalization ensures that a smaller divergence (indicating better alignment with the teacher) yields a higher reward value. The total reward is then computed as the sum of the accuracy and distillation rewards.

\vspace{1.5mm}
\noindent\textbf{Final Objective.}
We adopt RL objective of GRPO~\cite{shao2024deepseekmath} (see Appendix~\ref{sec:appB}) and extend it with \cref{eq:5}, as follows:
\begin{equation}
\footnotesize
\min_{\mathbf{W}_\mathcal{S}}\mathbb{E}_{(x,\hat{y})\sim \text{Gen-Data}[i]}\left[\mathcal{L}_{\text{GRPO}} + \mathcal{D}\left( 
P_{\mathcal{T}_{r[i]}}(\hat{y}|x) \, \Vert \, P_{\mathcal{S}}(\hat{y}|x)\right)\right].
\end{equation}
This unified objective integrates reinforcement and distillation, allowing the student to refine its responses in a stable manner. \cref{fig:overview} and \cref{fig:3} represents overview of \textbf{Masters}.

% table 4
\begin{table*}[t!]
\centering
\vspace{0mm}
\caption{Ablation studies on various configurations influencing the performance of \textbf{Masters}. Note that (a) reports the average performance across the evaluation benchmarks in \cref{tab:1}, while (c), (d), and (e) are conducted using the InternVL3.5 series~\cite{wang2025internvl3}, with 8B student.}
\label{tab:4}
\centering
\vspace{-3mm}
\begin{minipage}[t]{0.32\linewidth}
\caption*{(a) Maximum of Masking Ratio}
\centering
\vspace{-3mm}
\resizebox{\linewidth}{!}{
\renewcommand{\tabcolsep}{1mm}
\begin{tabular}{lcccc}
\toprule
$r_{\max}$   & Q2.5-VL-72B    & Q3-VL-32B    & IVL3-78B      & IVL3.5-38B      \\
\midrule
0   & 72.7 & 77.7 & 73.9 & 76.8 \\
0.1 & 75.9 & 78.3 & 75.1 & 78.1 \\
0.2 & \cellcolor{colorful}\textbf{79.4} & 79.0 & \cellcolor{colorful}\textbf{80.5} & \cellcolor{colorful}\textbf{80.0} \\
0.3 & 70.4 & 79.5 & 71.1 & 70.4 \\
0.4 & 64.5 & \cellcolor{colorful}\textbf{80.4} & 68.9 & 66.2 \\
0.5 & 48.2 & 75.4 & 50.0 & 49.2 \\
\bottomrule
\end{tabular}
}

\vspace{+1.2mm}
\caption*{(d) Decomposing Components}
\centering
\vspace{-3mm}
\resizebox{\linewidth}{!}{
\renewcommand{\tabcolsep}{2.5mm}
\begin{tabular}{lcccc}
\toprule
                 & Mid Teacher & Multi-Response & Reward Feedback & Avg \\
\midrule
Naive            & \xmark      & \xmark         & \xmark          & 75.8 \\
Naive            & \xmark      & \cmark         & \xmark          & 76.2 \\
Naive            & \xmark      & \cmark         & \cmark          & 76.8 \\
\rowcolor{colorful}
Naive            & \cmark      & \cmark         & \cmark          & \textbf{77.3} \\
\midrule
Mask-Progressive & \xmark      & \xmark         & \xmark          & 76.0 \\
Mask-Progressive & \xmark      & \cmark         & \xmark          & 76.3 \\
Mask-Progressive & \xmark      & \cmark         & \cmark          & 77.1 \\
\rowcolor{colorful}
Mask-Progressive & \cmark      & \cmark         & \cmark          & \textbf{80.0} \\
\bottomrule
\end{tabular}
}

\end{minipage}
\begin{minipage}[t]{0.32\linewidth}
\centering
\caption*{(b) Combinations of Teacher}
\vspace{-3mm}
\resizebox{\linewidth}{!}{
\renewcommand{\tabcolsep}{0.8mm}
\begin{tabular}{lccccc}
\toprule
Teacher Size & AI2D & MathVista & MMB & MM-Vet & MMMU \\
\midrule
38B          & 86.8 & 73.4 & 85.3 & 82.1 & 65.1 \\
78B          & 86.9 & 75.7 & 86.8 & 80.2 & 67.8 \\
14B+38B      & 87.2 & 76.8 & 87.1 & 82.5 & 69.0 \\
14B+78B      & 87.5 & 78.3 & 88.9 & 82.9 & 70.6 \\
38B+78B      & 88.1 & 80.8 & \textbf{90.1} & 83.0 & 72.9 \\
\rowcolor{colorful}
14B+38B+78B  & \textbf{88.9} & \textbf{82.3} & \textbf{90.1} & \textbf{83.8} & \textbf{74.0} \\
\bottomrule
\end{tabular}
}
\caption*{(e) Choice of Reward Design}
\centering
\vspace{-3mm}
\resizebox{\linewidth}{!}{
\renewcommand{\tabcolsep}{0.7mm}
\begin{tabular}{lccccc}
\toprule
Reward                                                                   & AI2D & MathVista & MMB & MM-Vet & MMStar \\
\midrule
$\mathcal{R}_{\text{acc}} (\triangle)$                                   & 82.3 & 75.6 & 80.7 & 77.2 & 65.3 \\
$\mathcal{R}_{\text{acc}}$                                               & 86.5 & 82.3 & 88.1 & 85.0 & 75.8 \\
$\mathcal{R}_{\text{distill}}$                                           & 86.3 & 80.3 & 88.0 & 84.8 & 72.0 \\
$\mathcal{R}_{\text{acc}}+\mathcal{R}_{\text{distill}}(\triangle)$       & 86.6 & 82.5 & 88.1 & 85.2 & 76.3 \\
\rowcolor{colorful}
$\mathcal{R}_{\text{acc}}+\mathcal{R}_{\text{distill}}$                  & 87.2 & \textbf{85.0} & 88.2 & \textbf{85.6} & \textbf{80.8} \\
InternVL3.5-38B                                                          & \textbf{87.8} & 81.9 & \textbf{90.3} & 82.2 & 75.3 \\
\bottomrule
\end{tabular}
}
\end{minipage}
\begin{minipage}[t]{0.32\linewidth}
\caption*{(c) Source of Training Samples}
\centering
\vspace{-3mm}
\resizebox{\linewidth}{!}{
\renewcommand{\tabcolsep}{1.4mm}
\begin{tabular}{lccccc}
\toprule
Source                                & AI2D & MathVsita & MMB & MM-Vet & MMMU \\
\midrule
$\mathcal{T}(\#8)$                    & 87.0 & 84.8 & 86.3 & 85.3 & 72.6 \\
$\mathcal{S}(\#2) + \mathcal{T}(\#6)$ & 87.0 & 84.6 & 86.8 & 85.0 & 72.4 \\
$\mathcal{S}(\#3) + \mathcal{T}(\#5)$ & 87.1 & 84.8 & 87.2 & 85.2 & 72.2 \\
\rowcolor{colorful}
$\mathcal{S}(\#4) + \mathcal{T}(\#4)$ & \textbf{87.2} & \textbf{85.0} & \textbf{88.2} & \textbf{85.6} & \textbf{72.7} \\
$\mathcal{S}(\#5) + \mathcal{T}(\#3)$ & 87.0 & 84.6 & 87.0 & 85.1 & 72.0 \\
$\mathcal{S}(\#6) + \mathcal{T}(\#2)$ & 86.8 & 84.3 & 86.5 & 84.6 & 71.5 \\
$\mathcal{S}(\#8)$                    & 86.4 & 83.8 & 86.3 & 85.0 & 71.0 \\
\bottomrule
\end{tabular}
}

\caption*{(f) Other Distillation Methods}
\vspace{-3mm}
\resizebox{\linewidth}{!}{
\renewcommand{\tabcolsep}{4.5mm}
\begin{tabular}{lccccc}
\toprule
VLMs                                              & \textbf{Masters} & MathVista & MMB & MM-Vet & MMMU \\
\midrule
\multirow{2}{*}{DistiLLM-7B~\cite{ko2024distillm}}& \xmark           & 61.5      & 84.8& 65.3   & 57.9 \\
\cdashline{2-6}\noalign{\vskip 0.5ex}
                                                  & \cellcolor{colorful}\cmark           & \cellcolor{colorful}\textbf{69.3}      & \cellcolor{colorful}\textbf{86.0}& \cellcolor{colorful}\textbf{73.3}   & \cellcolor{colorful}\textbf{60.8} \\
\midrule
\multirow{2}{*}{LLaVA-KD-7B~\cite{cai2024llava}}  & \xmark           & 61.8    & 85.0& 65.9   & 58.2 \\
\cdashline{2-6}\noalign{\vskip 0.5ex}
                                                  & \cellcolor{colorful}\cmark           & \cellcolor{colorful}\textbf{70.2}    & \cellcolor{colorful}\textbf{86.2}& \cellcolor{colorful}\textbf{73.8}   & \cellcolor{colorful}\textbf{61.7} \\
\midrule
\multirow{2}{*}{VLsI-7B~\cite{lee2024vlsi}}       & \xmark           & 74.7     & 85.8&  75.2  & 69.3 \\
\cdashline{2-6}\noalign{\vskip 0.5ex}
                                                  & \cellcolor{colorful}\cmark           & \cellcolor{colorful}\textbf{75.1}     & \cellcolor{colorful}\textbf{86.9}& \cellcolor{colorful}\textbf{75.8}   & \cellcolor{colorful}\textbf{69.8} \\
\midrule
\multirow{2}{*}{RIL-8B~\cite{lee2025unified}}     & \xmark           & 77.8 &  88.1  & 80.1   & 68.6 \\
\cdashline{2-6}\noalign{\vskip 0.5ex}
                                                  & \cellcolor{colorful}\cmark           & \cellcolor{colorful}\textbf{82.3} & \cellcolor{colorful}\textbf{90.1}   & \cellcolor{colorful}\textbf{83.8}   & \cellcolor{colorful}\textbf{74.0}  \\
\bottomrule
\end{tabular}
}
\end{minipage}
\vspace{0mm}
\end{table*}

\section{Experiments}
\label{sec:experi}

\subsection{Implementation Details}
\label{sec:4.1}
\vspace{2mm}
\noindent\textbf{Model Selection.} We select recently released, high-performing student and teacher VLMs based on Leaderboard~\cite{2023opencompass}. For student, we employ several strong series, including Qwen2.5-VL-3B and -7B~\cite{bai2025qwen2}, Qwen3-VL-2B, -4B, and -8B, InternVL3-2B, and -8B~\cite{zhu2025internvl3}, as well as InternVL3.5-2B, -4B, and -8B~\cite{wang2025internvl3}. For teacher, we select corresponding family of Qwen2.5-VL-32B and -72B~\cite{bai2025qwen2}, Qwen3-VL-32B~\cite{yang2025qwen3}, InternVL3-14B, -38B, and -78B~\cite{zhu2025internvl3}, and InternVL3.5-14B and -38B~\cite{wang2025internvl3}.

\vspace{2mm}
\noindent\textbf{Training Setup.} We train and evaluate \textbf{Masters} primarily on NVIDIA A100 80GB GPUs. We first set the decrement $s$ to $0.05$ and save multiple masked teacher checkpoints with different masking ratios. For instance, when $r_{\max}$ is set to $0.2$, we store five masked teacher checkpoints corresponding to masking ratios of $0.20$, $0.15$, $0.10$, $0.05$, and $0$. We then make all five masked teachers generate responses for the 1.5M dataset (see Appendix~\ref{sec:appC}), by using \texttt{vLLM}~\cite{kwon2023efficient} for fast inference. Specifically, we generate $8$ responses per question by setting the temperature to $1.0$, top-p to $0.9$, top-k to $50$, and the repetition penalty to $1.05$. We simultaneously evaluate the accuracy reward of the responses via LLM-as-a-Judge~\cite{zheng2023judging}. The model used for judging is the same as the one selected for generation, and we further refine the accuracy reward through additional parsing prompts (see Appendix~\ref{sec:appD}). For RL, we utilize the DeepSpeed engine with ZeRO-3~\cite{rajbhandari2020zero} to efficiently handle large teacher and student. The student is optimized using AdamW~\cite{loshchilov2018decoupled} with a fixed learning rate of $1\times10^{-6}$.

\subsection{Step-wise Dissection of Masters}
We conduct a series of step-wise experiments to construct \textbf{Masters} and identify the source of its effectiveness. The results of each distillation strategy are summarized in \cref{tab:1}. We begin with a naive baseline that performs distillation from a large teacher using the objective of JSD~\cite{xu2024llavadi, agarwal2024policy}, without incorporating mask-progressive distillation or RL feedback. Next, we apply teacher masking and perform mask-progressive distillation based on \cref{eq:5}, which yields clear improvements—confirming that capacity alignment between the teacher and the student stabilizes optimization. Finally, adding RL allows the student to further refine the responses through feedback (correctness and transferability). Overall, combining mask-progressive distillation with RL feedback results in a cumulative performance boost, demonstrating that each component contributes synergistically to the overall effectiveness of \textbf{Masters}.

We further analyze the impact of gradually increasing the teacher sizes during distillation (e.g., from 14B to 38B), as shown in \cref{tab:2}. We first distill knowledge from an intermediate-sized (mid) teacher as a warm-up stage, and then perform distillation from a larger teacher. This gradual teacher-size scaling ({Blue Color}) consistently yields better performance than one-shot distillation from a large teacher (Green Color), indicating that progressive capacity alignment leads to smoother convergence and richer representations. \cref{fig:2}(a) illustrates the masking ratio schedule and performance trends across different teacher settings. Notably, after the warm-up stage with the mid teacher shown in \cref{fig:2}(b), performance accelerates significantly when distilling from the large teacher, highlighting the advantage of this teacher scaling strategy. In summary, when combined with mask-progressive distillation and RL feedback, \textbf{Masters} elevates the student to a highly competitive level—matching or even surpassing many recent open- and closed-source VLMs, as reported in \cref{tab:3} and \cref{tab:5} across various model scales (\cref{fig:1}).

% figure 5
\begin{figure}[t!]
\vspace{0mm}
    \centering
    \includegraphics[width=0.35\textwidth]{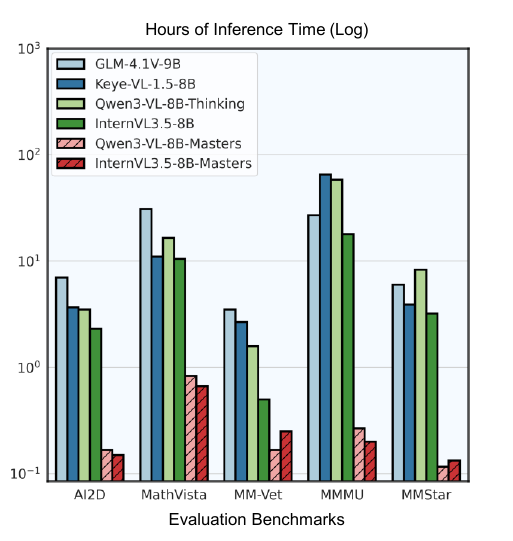}
    \vspace{-3mm}
    \caption{Comparing hours of inference time at one GPU across various models described in \cref{tab:3}}
    \label{fig:5}
    \vspace{0mm}
\end{figure}

\subsection{Ablation Studies with Configurations}
We conduct comprehensive ablation studies to examine various configuration settings that influence the detailed performance of \textbf{Masters}. First, we investigate the number of \textit{pre-generated} responses in \cref{fig:4}. As this number increases, performance steadily improves and converges around eight responses. Therefore, we set the number of generated responses to eight to achieve a great balance between training efficiency and performance.

Next, \cref{tab:4}(a) shows a grid search to determine the optimal maximum masking ratio ($r_{\max}$) for each teacher, ranging from $0$ to $0.5$. For most teachers, the optimal value is $r_{\max}=0.2$, whereas Qwen3-VL-32B~\cite{yang2025qwen3} achieves its best results with $r_{\max}=0.4$. Beyond this table, we similarly observe that InternVL3.5-8B and -14B~\cite{wang2025internvl3} also perform best at $r_{\max}=0.4$. We then analyze the effect of teacher composition when distilling InternVL3 series~\cite{zhu2025internvl3} from large teachers to an 8B student. Consistent with the results in \cref{tab:2}, the best performance is achieved with gradual teacher-size scaling of 14B, 38B, and 78B in \cref{tab:4}(b). This confirms that progressive teacher scaling enables more stable and capacity-aligned knowledge transfer.

Furthermore, we examine how the source of generated responses affects performance. As shown in \cref{tab:4}(c), using only teacher-generated responses limits the student’s adaptability to its own representational capacity, while using only student-generated responses restricts exposure to the teacher’s richer semantics. Consequently, we adopt a balanced 1:1 ratio of teacher- and student-generated responses to ensure both adaptability and semantic richness. \cref{tab:4}(d) compares various configurations, including naive versus mask-progressive distillation, with and without mid-teacher scaling, multi-response generation, and reward feedback (\cref{fig:2}(c)). The results demonstrate that each component contributes synergistically to performance, and removing any of them leads to degradation—highlighting their essential roles in performance acceleration.

Next, we analyze the impact of reward design by selectively removing the accuracy or distillation reward. The triangle symbol ($\triangle$) for the accuracy reward indicates the rule-based evaluation, which does not rely on LLM-as-a-Judge~\cite{zheng2023judging}, while the triangle for the distillation reward denotes exclusion of reverse min–max normalization. Removing the accuracy reward consistently results in performance below that of the large teacher, underscoring its importance in enabling the student to surpass the teacher. Without normalization, the distillation reward becomes less discriminative, yielding performance comparable to using only the accuracy reward. As shown in \cref{tab:4}(e), both accuracy and distillation rewards are essential for effectively identifying high-quality responses.

Finally, we apply \textbf{Masters} to other recent distillation frameworks: DistiLLM~\cite{ko2024distillm}, LLaVA-KD~\cite{cai2024llava}, VLsI~\cite{lee2024vlsi}, and RIL~\cite{lee2025unified}. \textbf{Masters} consistently mitigates degradation caused by large parameter gaps. Notably, while most existing methods rely on multi-step pipelines with additional SFT phases to maintain stability, \textbf{Masters} achieves superior results in a single-step training process—demonstrating its simplicity, scalability, and training efficiency.

\section{Discussion and Conclusion}
\label{sec:conclu}

Although \textbf{Masters} shows strong performance and scalability, it is trained in an offline manner, which limits adaptability to real-time feedback. In principle, training could continuously improve the model via online data sampling, but it remains computationally infeasible—for instance, even a single model requires over 30 days on 256 A100 GPUs for 1.5M samples, whereas \textbf{Masters} completes training in just two days. Future advances in efficient multi-node training and inference from \texttt{vLLM}-like libraries is strongly needed for online distillation within days.

Beyond this limitation, recent methods often adopt a ``think-answer'' paradigm, which improves reasoning but greatly increases latency (\cref{fig:5}). Such approaches remain impractical for real-world or on-device use due to their heavy computational cost. In contrast, \textbf{Masters} attains strong performance without sacrificing inference speed, suggesting a promising direction for balancing intelligence and scalability toward efficient VLMs.

We present \textbf{Masters}, progressively restoring the teacher from mask and reinforcing the student with dual rewards in an offline setting. It achieves stable and scalable knowledge transfer from large to compact models. Future work may explore data-driven masking and RL-based replay buffers to better handle hard examples. We hope \textbf{Masters} will inspire further research toward deployable and efficient VLMs.

\clearpage
{
  \small
  \bibliographystyle{unsrt}
  \bibliography{main}
}
\clearpage
\setcounter{page}{1}
% \maketitlesupplementary
\onecolumn

% Section A B C...
\appendix

\section{Related Work of Efficient VLMs}
\label{sec:appA}
With the advent of visual instruction tuning~\cite{liu2023visual, dai2023instructblip} and the scaling of large language models (LLMs)~\cite{kaplan2020scaling, chung2022scaling, chowdhery2023palm}, both large-scale open-source~\cite{bai2025qwen2, yang2025qwen3, zhu2025internvl3, wang2025internvl3} and closed-source~\cite{hurst2024gpt, comanici2025gemini, claude3series2024} vision–language models (VLMs) have emerged. However, these large-scale VLMs impose substantial computational demands in real-world scenarios~\cite{lee2025multiverse, lee2025refinebench}, such as on-device or edge processing. Consequently, there is a growing demand for lightweight VLMs that can be efficiently deployed on resource-constrained devices while maintaining fast inference, driving active research in efficient VLM design. Early efforts have mainly focused on integrating additional visual encoders~\cite{fang2023eva, oquab2023dinov2, kirillov2023segment, zhai2023sigmoid}, multiple computer vision backbones~\cite{cheng2022masked, minderer2023scaling, yang2022panoptic, du2021pp}, or rational embeddings~\cite{lee2024meteor, hao2024training, xu2025harnessing} into LLMs~\cite{lu2024deepseek, goncharova2024omnifusion, zong2024mova, lee2024collavo, lee2024moai, shi2024eagle}. In addition, a growing body of research~\cite{thawakar2024mobillama, mehta2024openelm, liu2024mobilellm, lee2024trol, lee2024phantom} has explored architectural strategies—such as shared or repetitive feed-forward network (FFN) structures and expanded hidden dimensions—to enhance efficiency without significant performance degradation. Furthermore, several studies~\cite{zhou2024tinyllava, chu2023mobilevlm, chu2024mobilevlm, lin2024moe, qiao2024vl, zhao2024cobra} propose vision–text aligned training strategies, adopt Mamba architecture~\cite{gu2023mamba}, or incorporate the mixture-of-experts paradigm~\cite{jacobs1991adaptive, eigen2013learning, bengio2013estimating, shazeer2017, riquelme2021scaling} to achieve scalable model capacity.

\section{The Objective of GRPO}
\label{sec:appB}

For a question \(x\) and its multiple generated responses \(\{\hat{y}_j\}_{j=1}^{G}\), the RL objective of GRPO~\cite{shao2024deepseekmath} (Generalized Reinforcement Policy Optimization) is defined as:
\begin{align}
&\mathcal{L}_{\text{GRPO}} 
= - 
\mathbb{E}_{j}\left[\mathbb{E}_{t}
\left[
\min \big( r_{j,t} A_{j}, \,
\mathrm{clip}(r_{j,t}, 1-\varepsilon, 1+\varepsilon) A_{j} \big)
- \beta D_{\mathrm{KL}}(\pi_\theta \| \pi_{\mathrm{ref}})
\right]\right],
\\[8pt]
&\quad\text{where} \quad
r_{j,t}
= \frac{\pi_\theta(\hat{y}_{j,t} \mid x, \hat{y}_{j,<t})}{\pi_{\theta_{\mathrm{old}}}(\hat{y}_{j,t} \mid x, \hat{y}_{j,<t})} \quad\text{and}\quad A_{j} = 
\frac{\mathcal{R}_j - \mathrm{mean}\left(\{\mathcal{R}_j\}_{j=1}^{G}\right)}
{\mathrm{std}\left(\{\mathcal{R}_j\}_{j=1}^{G}\right)}.
\end{align}
Here, $r_{j,t}$ denotes the policy ratio for new policy $\pi_{\theta}$ and old policy $\pi_{\theta_{\text{old}}}$ for each token $t$, and $A_{j}$ indicates the advantage computed by normalized rewards $\mathcal{R}$. This objective encourages the new policy $\pi_\theta$ to improve upon the old policy $\pi_{\theta_{\mathrm{old}}}$ according the advantage $A$. The clipped surrogate objective limits the policy update ratio $r_{j,t}$ to the range $[1-\varepsilon,\, 1+\varepsilon]$, preventing excessively large updates. In addition, KL divergence term $D_{\mathrm{KL}}(\pi_\theta \| \pi_{\mathrm{ref}})$ penalizes deviation from a reference policy $\pi_{\mathrm{ref}}$, ensuring regularization for stable training.

\noindent In our \textbf{Masters} training setup, the total reward is computed as the sum 
\(\mathcal{R} = \mathcal{R}_{\text{acc}} + \mathcal{R}_{\text{distill}}\), 
from which the advantage is directly derived. 
Since the updating model is the student, the policy \(\pi_\theta\) corresponds to the student's logit-softmax output \(P_{\mathcal{S}}\), 
and the parameter \(\theta\) represents the student's weight set \(\mathbf{W}_{\mathcal{S}}\). 
In our setup, the policy ratio \(r_{j,t}\) is always one because the student is updated only once per training iteration \(i\); hence, 
the old policy \(\pi_{\theta_{\mathrm{old}}}\) and the new policy \(\pi_{\theta}\) are identical. 
Therefore, the clipped surrogate term becomes redundant, and the objective of GRPO simplifies to
\begin{equation}
\mathcal{L}_{\text{GRPO}} 
= - \mathbb{E}_{j}\left[
r_{j,t}A_j
- \beta D_{\mathrm{KL}}(\pi_\theta \| \pi_{\mathrm{ref}})
\right],
\end{equation}
where \(r_{j,t}=1\). Technically, we still keep the ratio term in the expression to ensure the gradient properly flows to the student parameters during training. Additionally, we set \(\beta = 0.1\) to prevent the student from being updated excessively, providing stable regularization.

\clearpage
\section{Visual Instruction Tuning Data}
\label{sec:appC}

We assemble a 1.5M-sample visual instruction tuning dataset that encompasses both real-world and synthetic sources: COCO-ReM~\cite{singh2024benchmarking}, iNaturalist2018~\cite{van2018inaturalist}, VQA-v2~\cite{balanced_vqa_v2}, Super-CLEVR~\cite{li2023super}, MAVIS~\cite{zhang2024mavis}, Geometry3K~\cite{lu2021inter},  SQA~\cite{lu2022learn}, AI2D~\cite{kembhavi2016diagram}, SA-1B~\cite{kirillov2023segment}, LLaVAR~\cite{zhang2023llavar}, VSR~\cite{VSR}, TallyQA~\cite{acharya2019tallyqa}, TabMWP~\cite{lu2022dynamic}, KonIQ~\cite{hosu2020koniq}, InternVL~\cite{wang2024enhancing}-filtered synthetic knowledge dataset covering politics, math, physics, chemistry, RLAI-F~\cite{yu2024rlaif}, CLEVR-Math~\cite{lindstrom2022clevr}, SROIE~\cite{huang2019icdar2019}, ChartQA~\cite{masry2022chartqa}, DocVQA~\cite{mathew2021docvqa}, FigureQA~\cite{kahou2017figureqa}, GQA~\cite{hudson2019gqa}, InfoVQA~\cite{mathew2022infographicvqa}, M3CoT~\cite{chen-etal-2024-m3cot}, MapQA~\cite{chang2022mapqa}, OK-VQA~\cite{marino2019ok}, TextVQA~\cite{singh2019towards}, WildVision~\cite{lu2024wildvision}, DVQA~\cite{kafle2018dvqa}, GeoQA+~\cite{cao2022augmented}, GeOS~\cite{seo2015solving}, IconQA~\cite{lu2021iconqa}, UniGEO~\cite{chen2022unigeo}, GeomVerse~\cite{kazemi2023geomverse}, Geo170K~\cite{gao2023g}, MathV360K~\cite{shi2024math}, and RAM++~\cite{huang2023open}-filtered synthetic data of Infinity-MM~\cite{gu2024infinity} covering coarse and fine-grained perception, relation, attribute, and logic reasoning.

\section{Additional Parsing Prompts for Accuracy Reward}
\label{sec:appD}
\begin{tcolorbox}[colback=black!5!white, colframe=black!70!white, title=Prediction Evaluation Prompt, fonttitle=\bfseries]
\begin{verbatim}
System:
You are an evaluation assistant that gives accuracy scores
compared with Ground Truth and Generated Text from AI.
Question is in <question> </question> tag.
Ground Truth is in <ground truth> </ground truth> tag.
Generated Text in <generated text> </generated text> tag.
After reading the Question, compare the Generated Text
against the Ground Truth summary:
  - If the Generated Text fully and correctly captures the core point → 1
  - If it is incorrect or irrelevant → 0
  - If it has repetitive response → 0
  - If it has empty response → 0
Output the numerical evaluation score (0 or 1) after giving a brief explanation.
  - The evaluation score should be wrapped in <answer> </answer> tag.

User:
<question>
{}
</question>

<ground truth>
{}
</ground truth>

<generated text>
{}
</generated text>

Provide the numerical evaluation score after giving a brief explanation.
The evaluation score should be wrapped in <answer> </answer> tag.
\end{verbatim}
\end{tcolorbox}

\begin{tcolorbox}[colback=black!5!white, colframe=black!70!white, title=Accuracy Reward Parsing Prompt, fonttitle=\bfseries]
\begin{verbatim}
System:
You are an evaluation assistant that gives binary accuracy scores (0 or 1)
based on the provided overall summary.
The summary will be wrapped inside <overall_summary>
and </overall_summary> tag.
After reading the summary, briefly output the integer score (0 or 1)
without any text.
Your final output must include only the integer value.

User:
<overall_summary>
{}
</overall_summary>

Please output your integer accuracy score (0 or 1)
based on the summary above without any text.
\end{verbatim}
\end{tcolorbox}

\end{document}